%% file: DiscoveryBench.tex
\documentclass{article}

\usepackage[preprint,nonatbib]{neurips_data_2024}
\usepackage[numbers]{natbib}
\usepackage{algorithm}
\usepackage{algpseudocode}
\usepackage{amsmath}

\usepackage[utf8]{inputenc} 
\usepackage[T1]{fontenc}    
\usepackage{hyperref}       
\usepackage{url}            
\usepackage{booktabs}       
\usepackage{amsfonts}       
\usepackage{nicefrac}       
\usepackage{microtype}      
\usepackage{xcolor}         

\usepackage{enumitem}
\usepackage{graphicx}
\usepackage{subfigure}
\usepackage{wrapfig}

\input{macros}

\input{utils}
\usepackage{cleveref}

\title{\textsc{DiscoveryBench}: Towards Data-Driven Discovery with Large Language Models}

\author{%
    Bodhisattwa Prasad Majumder$^{*1}$~~~ Harshit Surana$^{*12}$~~~ Dhruv Agarwal$^{*3}$\\\textbf{Bhavana Dalvi Mishra$^{*1}$~~~ Abhijeetsingh Meena$^{2}$~~~ Aryan Prakhar$^{2}$~~~ Tirth Vora$^{2}$} \\\textbf{Tushar Khot$^{1}$~~~ Ashish Sabharwal$^{1}$~~~ Peter Clark$^{1}$} \\
    $^{1}$Allen Institute for AI ~~~ $^{2}$OpenLocus ~~~ $^{3}$University of Massachusetts Amherst \\
    \\
    Website: \url{https://github.com/allenai/discoverybench}\\
    \hf{}~ \url{https://huggingface.co/datasets/allenai/discoverybench}\\
    $^{*}$equal contributions
}

\begin{document}

\maketitle

\input{sections/0_abstract}
\input{sections/1_introduction}
\input{sections/6_related_work}

\input{sections/2_background}
\input{sections/3_discovery_bench}
\input{sections/5_experiments}
\input{sections/7_conclusion}

\bibliographystyle{abbrvnat}
\bibliography{custom}

\input{sections/8_appendix}

\end{document}

%% file: macros.tex
\usepackage{enumitem}
\usepackage{tikz}
\usepackage{pgfplots}
\usepackage{xspace}
\usepackage[frozencache,cachedir=.]{minted}

\usepackage{xcolor} 
\definecolor{LightGray}{gray}{0.7}

\usepackage[T1]{fontenc}

\usepackage[utf8]{inputenc}

\usepackage{microtype}
\usepackage{makecell}

\usepackage{array}
\usepackage{booktabs}
\usepackage{soul}
\usepackage{url}
\usepackage[utf8]{inputenc}
\usepackage{graphicx}
\usepackage{amsmath}
\usepackage{booktabs}
\usepackage{colortbl}
\usepackage{multirow}
\usepackage{float}	
\usepackage{dblfloatfix}
\usepackage{setspace}  

\usepackage{xcolor}	

\usepackage{cleveref}

\definecolor{ForestGreen}{rgb}{0.13, 0.55, 0.13}
\definecolor{skyblue}{HTML}{46C5DD}	

\usepackage{quoting}

\usepackage[]{mdframed}
\usepackage{framed}

\usepackage{ctable} 
\usepackage{multicol}

\renewcommand{\cite}{\citep}

\definecolor{lightgray}{gray}{0.9}
\definecolor{Box1Color}{RGB}{227, 236, 246}
\definecolor{Box2Color}{RGB}{248, 220, 225}
\definecolor{Box3Color}{RGB}{255, 238, 224}
\definecolor{cbBlue}{RGB}{0, 114, 178}
\definecolor{cbOrange}{RGB}{240, 228, 66}
\definecolor{cbGreen}{RGB}{0, 158, 115}
\definecolor{cbRed}{RGB}{213, 94, 0}
\definecolor{cbPurple}{RGB}{204, 121, 167}
\definecolor{cbSkyBlue}{RGB}{86, 180, 233}
\definecolor{cbGray}{RGB}{128, 128, 128}
\definecolor{CBF1}{RGB}{255,99,132}  
\definecolor{CBF2}{RGB}{54,162,235}  
\definecolor{CBF3}{RGB}{255,206,86}  
\definecolor{CBF4}{RGB}{75,192,192}  
\definecolor{CBF5}{RGB}{153,102,255} 
\definecolor{CBF1b}{RGB}{205,89,112}  
\definecolor{CBF2b}{RGB}{44,142,215}  
\definecolor{CBF5b}{RGB}{133,92,225}  
\definecolor{PromptBgColor}{RGB}{255, 238, 224}

\usepackage{amssymb}
\usepackage{pifont}

%% file: utils.tex
\newcount\Comments  
\definecolor{darkgreen}{rgb}{0,0.5,0}
\definecolor{darkred}{rgb}{0.7,0,0}
\definecolor{teal}{rgb}{0.1,0.6,0.7}
\definecolor{blue}{rgb}{0.0,0.1,0.9}
\definecolor{orange}{rgb}{1.,0.7,0.0}
\definecolor{lightblue}{rgb}{0.70, 0.80, 0.89}
\definecolor{violet}{rgb}{0.50, 0.16, 0.88}
\newcommand{\kibitz}[2]{\ifnum\Comments=1{{\textcolor{#1}{\textsf{\footnotesize [#2]}}}}\fi}
\newcommand{\dhruv}[1]{\kibitz{darkred}{Dhruv: #1}}

\newcommand{\ashish}[1]{\kibitz{darkgreen}{Ashish: #1}}

\newcommand{\eat}[1]{}

\newcommand{\discoverybench}{\textsc{DiscoveryBench}}
\newcommand{\real}{\textsc{DB-Real}}
\newcommand{\synth}{\textsc{DB-Synth}}

\newenvironment{ite}{                     
     \parskip 0cm \begin{itemize}[leftmargin=*] \parskip 0cm \parsep 0cm \itemsep 0cm \topsep 0cm}{
        \end{itemize}} 

\newcommand{\newpara}[1]{\textbf{#1} \hspace{0.3em}}

\DeclareRobustCommand{\hf}{%
  \begingroup\normalfont
  \includegraphics[height=1.2\fontcharht\font`\M]{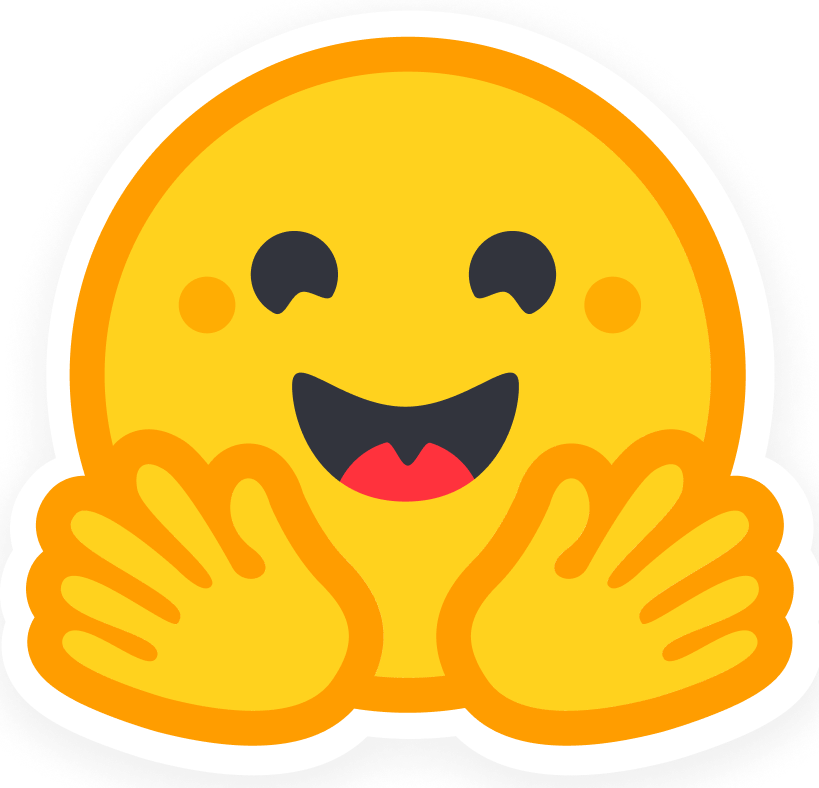}%
  \endgroup
}

%% file: sections/0_abstract.tex
\begin{abstract}
Can the rapid advances in code generation, function calling, and data analysis using large language models (LLMs) help automate the search and verification of hypotheses purely from a set of provided datasets?
To evaluate this question, we present \discoverybench{}, the first comprehensive benchmark that formalizes the multi-step process of data-driven discovery.
The benchmark is designed to systematically assess current model capabilities in discovery tasks and provide a useful resource for improving them.
Our benchmark contains 264 tasks collected across 6 diverse domains, such as sociology and engineering, by manually deriving discovery workflows from published papers
to approximate the real-world challenges faced by researchers, where each task is defined by a dataset, its metadata, and a discovery goal in natural language.
We additionally provide 903 synthetic tasks to conduct controlled evaluations across task complexity.
Furthermore, our structured formalism of data-driven discovery enables a facet-based evaluation that provides useful insights into different failure modes. 
We evaluate several popular LLM-based reasoning frameworks using both open and closed
LLMs as baselines on \discoverybench{} and find that even the best system scores only 25\%. 
Our benchmark, thus, illustrates the challenges in autonomous data-driven discovery and serves as a valuable resource for the community to make progress.
\end{abstract}


%% file: sections/1_introduction.tex
\section{Introduction}
\label{sec:introduction}

\eat{
\begin{figure}
    \centering
  \includegraphics[width=1\columnwidth]{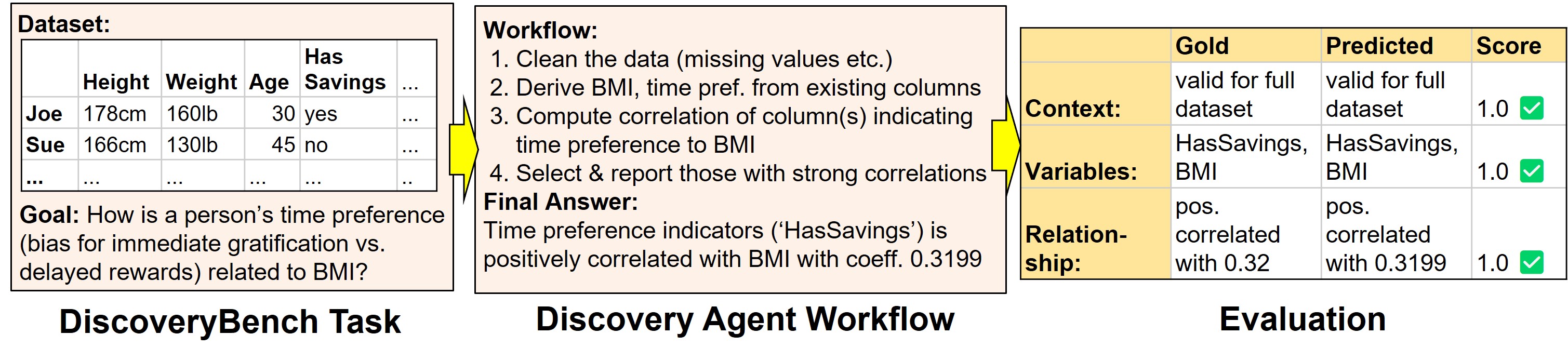}
    \caption{Each DiscoveryBench task consists of a goal and dataset(s) (left). Solving the task requires both statistical analysis and semantic reasoning, e.g., mapping goal terms to column names (center). A faceted evaluation allows open-ended final answers to be rigorously evaluated (right).}
    \label{fig:teaser}
\end{figure}
}

Knowledge discovery via the scientific process 
has been a catalyst for human progress for centuries
but has, thus far, been a predominantly manual pursuit \citep{glass2008brief}. 
Recent breakthroughs in capabilities of large language models (LLMs) to reason and interface with the world using code \citep{chen2021evaluating, roziere2023code}, external tools \citep{schick2024toolformer}, and interactive agents \citep{yao2023react, majumder2023clin}, however, now suggest the possibility of realizing a discovery system that is fully autonomous.
Indeed, recent works \citep{majumder2024data} provide initial evidence for this paradigm within the setting of \emph{data-driven discovery}, where both search and verification of hypotheses may be carried out using a dataset alone (i.e., 
after physical experiments and data collection\footnote{
    In practice, experiments and analysis are interleaved, not sequential. Our concern
    in this work, however, is systematically studying the data analysis part of the (interleaved) pipeline.
    }),
but the extent of this ability remains unclear. 
We, therefore, aim to systematically evaluate the following question: 

{\centering
    \emph{How good are current state-of-the-art LLMs at automated data-driven discovery?}\par
}

\begin{figure}
    \centering
  \includegraphics[width=1\columnwidth]{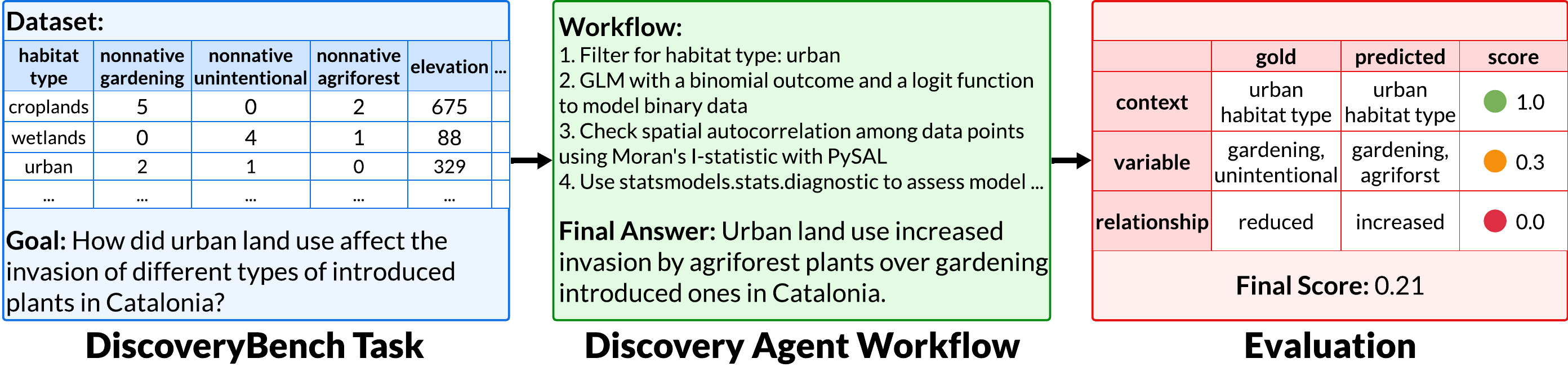}
 \caption{\small Each \discoverybench{} task consists of a goal and dataset(s) (left). Solving the task requires both statistical analysis and scientific semantic reasoning, e.g., deciding which analysis is appropriate for the domain, and mapping goal terms to column names (center). A faceted evaluation allows open-ended final answers to be rigorously evaluated (right).}
    \label{fig:teaser}
\end{figure}

Answering this question is hard, as data-driven discovery in the wild (real-world) is
diverse across domains and subject areas, which in turn makes it difficult to build
a robust evaluation framework to measure progress.
We address this using a pragmatic formalization of data-driven discovery, namely
the search for a {\it relationship} that may hold between {\it variables} in a {\it context},
where (importantly) the description of those facets may not be in the language of the
dataset. A data-driven discovery task then has one of these components missing, e.g., ``How
did urban land use affect the invasion of introduced plants in Catalonia?".
Importantly, this formalization allows for systematic,
reproducible evaluation over a wide variety of real-world problems, by leveraging these facets (Fig~\ref{fig:teaser}, right).

Unlike prior datasets for statistical analysis \cite{liu2024llms} or AutoML \cite{Zhang2023AutoMLGPTAM, Gijsbers2022AMLBAA}, 
\discoverybench{} tasks
also require scientific semantic reasoning, for instance, deciding which of the many possible
analysis techniques are appropriate for the domain (e.g., spatial autocorrelation for
plant invasion, Fig~\ref{fig:teaser} center), how to clean and/or normalize the data, 
and how to map goal terms to dataset terms (e.g., ``land use'' to ``habitat type''). Task
solutions typically requires a multistep workflow (Fig~\ref{fig:teaser}, center). 
In this way, \discoverybench{} is the first
large-scale dataset to address the broader data-driven discovery pipeline, not just
the statistical analysis component, and explore LLMs' capacity for this.

Given this framework, we created \discoverybench{} by manually extracting 264 discovery tasks, i.e., goal + dataset(s), from over 20 published papers, as well as creating real-world discovery workflows
that solve each task. We additionally provide 903 synthetic tasks across 48
domains generated using LLMs to mimic the real-world discovery process. The synthetic
benchmark allows us to conduct controlled model evaluations by varying task difficulty.
Our contributions are thus:
\begin{ite}
  \item \discoverybench, the first comprehensive benchmark to formalize 
    the multi-step process of data-driven hypothesis search and verification,
    covering many real-world discovery tasks plus additional synthetic tasks.
  \item A pragmatic formalism for data-driven discovery, flexible enough to
    characterize many real-world tasks while constrained enough to allow
    for rigorous, reproducible evaluation.
  \item A comprehensive evaluation across state-of-the-art LLM-based reasoning methods (``discovery agents'').
    We find performance peaks at 25\%, demonstrating the challenging nature of our task.
\end{ite}
These suggest that \discoverybench{} may be a valuable resource for helping make progress
on autonomous, data-driven discovery.


%% file: sections/6_related_work.tex
\section{Related Work}
\label{sec:related_work}

Automated data-driven discovery has been a long-standing dream of AI \cite{majumder2024data,kitano2016artificial}.
Although there have been a range of {\bf data-driven discovery systems},
from early ones that fit equations to idealized data, e.g., Bacon \citep{Langley1981DataDrivenDO},
to more modern ones handling complex real-world problems, e.g., AlphaFold \cite{jumper2021highly},
their associated datasets are task-specific and customized to a
pre-built pipeline. In contrast, \discoverybench{} aims to be a general
test over multiple tasks, including testing whether systems can
design appropriate pipelines themselves.

A number of datasets and tools are available for {\bf AutoML}, a related
technology aimed at automating workflows for building optimal machine
learning models \cite{Jin2023AutoKerasAA,Zhang2023AutoMLGPTAM,LeDell2020H2OAS}.
AutoML tools include packages like Scikit \cite{feurer2015efficient},
and embedded in cloud platforms such as Google Cloud Platform,
Microsoft Azure,
and Amazon Web Services.
However, associated datasets for AutoML are primarily used for training models,
rather than for open-ended discovery tasks.

Similarly, there are several datasets that test {\bf statistical
analysis} in various fields, e.g., \cite{Shao2023ExploringPI,Li2024AutomatedSM,Yang2022ASL}.
Software packages like Tableaux, SAS, and R also support users in that task.
However, these datasets and tools are designed specifically for data analysis,
while \discoverybench{} aims to automate the broader pipeline including ideation,
semantic reasoning, and pipeline design, where statistical analysis is just one component.

One recent dataset similar in spirit to ours is QRData \cite{liu2024llms}.
QRData also explores LLM capabilities but targets statistical/causal analysis for well-defined (mainly) textbook questions that have unique, (mainly) numeric gold answers.
In contrast, \discoverybench{} has no prescribed boundaries on statistical techniques to apply,
uses open-ended questions and answers, and complex tasks drawn from state-of-the-art published work.

%% file: sections/2_background.tex
\section{Formalization}
\label{sec:background}

We begin by formalizing what we mean by a data-driven hypothesis and how the structure of a complex hypothesis may be viewed as a hypothesis semantic tree.

A \textbf{data-driven hypothesis} $h$ in $\mathcal{H}$ (the space of such hypotheses) is a declarative sentence about the state of the world whose truth value may be inferred from a given dataset $D$ using a verification procedure $\mathcal{V}_D:\mathcal{H} \to \{\mathrm{supported}, \mathrm{unsupported}\}$, for instance, via statistical modeling. 

Each hypothesis may further be expressed using a propositional formula $\phi$ over a set of \textbf{sub-hypotheses} $h_i \in \mathcal{H}$ using logical connectives, e.g., disjunctions and conjunctions, such that $h := \phi(h_1, \ldots, h_n)$ and $\mathcal{V}_D(h)=\phi(\mathcal{V}_D(h_1),\ldots,\mathcal{V}_D(h_n))$. For instance, suppose $h$ is the hypothesis \textit{``for men younger than 20, popularity of product A varies proportional to their age \textbf{($h_1$)}, while there exists an inverse relationship for those older than 40 \textbf{($h_2$)}''}, then $h$ can be expressed as the conjunction $h_1 \wedge h_2$.

Inspired by recent work of \citet{thompson2023scope}, we additionally introduce a structured formalism that breaks a hypothesis down into \textbf{three hypothesis dimensions}:
\begin{ite}
    \item \textbf{Contexts $(c)$:} Boundary conditions that limit the scope of a hypothesis.
    E.g., \textit{``for men over the age of 30''} or \textit{``in Asia and Europe''} or unbounded/full dataset when not specified.
    \item \textbf{Variables $(v)$:} Known set of 
    concepts that interact in a meaningful way under a given context to produce the hypothesis. E.g., \texttt{gender}, \texttt{age}, or \texttt{income}. Note that each hypothesis is associated with a target variable and a set of independent variables.
    \item \textbf{Relationships $(r)$:} Interactions between a given set of variables under a given context that produces the hypothesis. E.g., \textit{``quadratic relationship''}, \textit{``inversely proportional''}, or piecewise conditionals.
\end{ite}
With slight abuse of notation, we can now equivalently define hypothesis $h:= \psi(c, v, r)$, where $\psi(\cdot,\cdot,\cdot)$ returns the declarative sentence
\textit{``under context $c$, variables $v$ have relationship $r$.''}
For instance, for sub-hypothesis $h_1$ in our example above, $c_1 :=$ \textit{``men younger than 20''}, $v_1 :=$ \{\texttt{gender}, \texttt{consumer\_age}, \texttt{product\_popularity}\}, and $r_1 := 
 $\textit{``popularity is proportional to age''}.


\newcommand{\hyptree}{\mathcal{T}_h}
\newcommand{\hypsubtree}{\mathcal{T}_{h'}}

\begin{wrapfigure}[16]{r}{0.45\textwidth}
  \centering
  \includegraphics[trim=0 0 0 0, clip, width=0.9\linewidth]{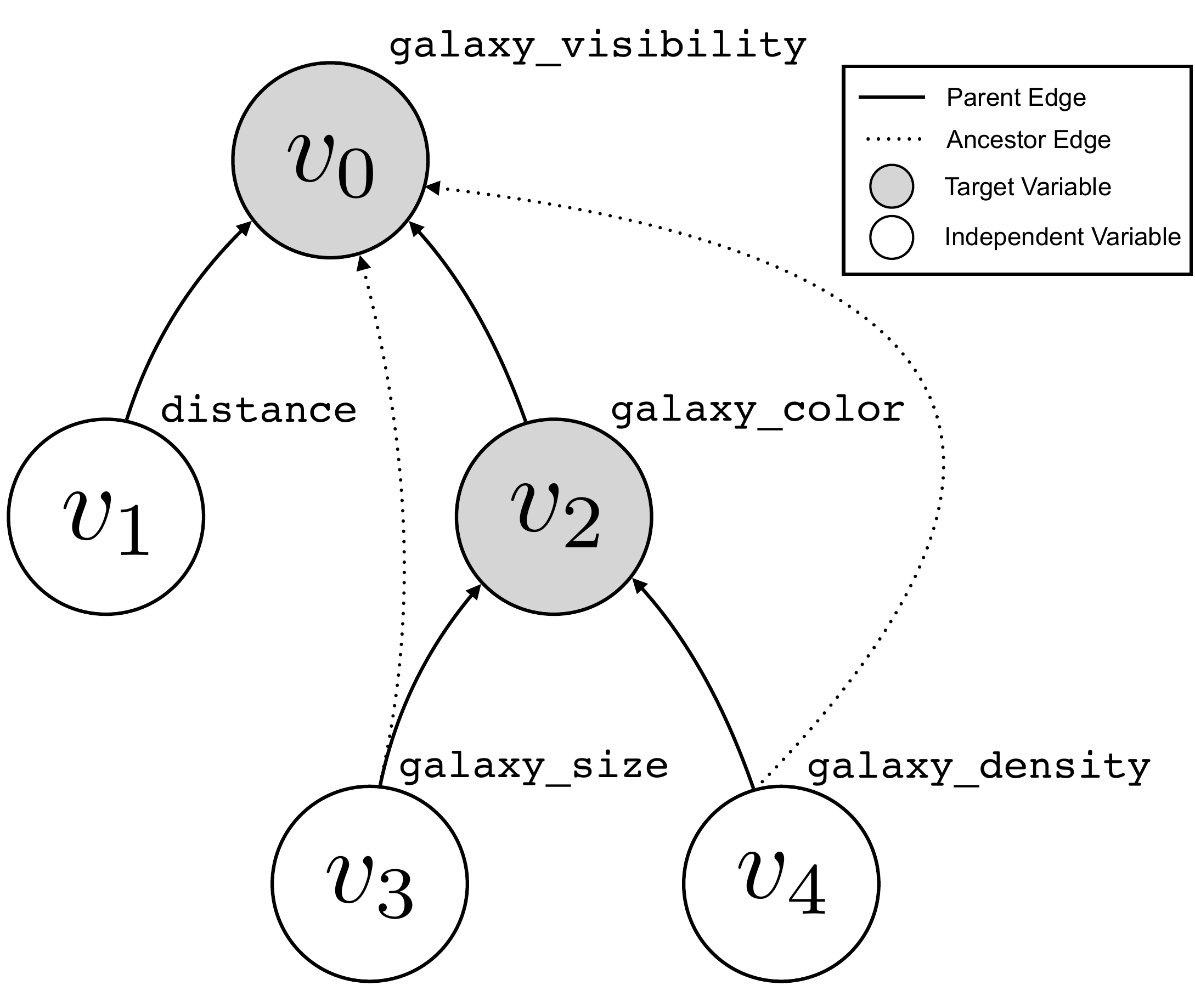}
  \caption{Hypothesis Semantic Tree}
  \label{fig:semantic_tree}
\end{wrapfigure}
\newpara{Hypothesis Semantic Tree.}
Observe that each independent variable in a hypothesis may itself be a target variable for a prior hypothesis.
To emphasize this hierarchical nature, we introduce the concept of a \emph{hypothesis semantic tree} whose nodes are variables (independent or derived) and whose sub-trees represent hypotheses, as follows. Consider a hypothesis $h$. 
A semantic hypothesis tree $\hyptree$ with $h$ as the \emph{primary} hypothesis is a Markov tree whose root node is the target variable of $h$, each of whose leaf nodes is an independent variable that is not derived further, and each of whose internal nodes is the target variable of an \emph{intermediate} hypothesis.
In other words, each sub-tree $\hypsubtree$ rooted at an internal node $v$ of $\hyptree$ is itself a hypothesis semantic tree for a hypothesis $h'$ with $v$ as the target variable. 
In particular, a sub-tree rooted at $v$ and all its immediate children $C_v$ implicitly encodes $h'$ as $\psi(c, \{v\} \cup C_v, r)$ where $r$ is the relationship between $v$ and $C_v$
under context $c$ as specified in $h'$.
More generally, $\hyptree$ can encode many different hypotheses by choosing one node as the target variable and considering nodes at arbitrary descendant levels as independent variables.

For instance, in Fig~\ref{fig:semantic_tree}, we show a semantic tree $\hyptree$ with the following primary hypothesis ($h$): \textit{``The visibility of a galaxy reduces when the blue spectrum dominates and the distance of the galaxy from Earth increases''}, where \texttt{galaxy\_visibility} ($v_0$) is the target variable with independent variables \texttt{distance} ($v_1$) and \texttt{galaxy\_color} ($v_2$). Consider now the sub-tree rooted at $v_2$, which encodes the following intermediate hypothesis: \textit{``Visibility of blue light from galaxies increases with an increase in galaxy size and decrease in star density''}, where \texttt{galaxy\_color} ($v_2$) is the target variable with independent variables \texttt{galaxy\_size} ($v_3$) and \texttt{galaxy\_density} ($v_4$). Note further that due to there existing ancestor edges from $v_3$ and $v_4$ to $v_0$, $\hyptree$ also encodes the hypothesis: \textit{``The visibility of a galaxy reduces with distance from Earth combined with an increase in galaxy size and decrease in star density''}, where the target variable is $v_0$ and the independent variables are $v_3$ and $v_4$.

\newpara{(Task) Dataset.} We formally define a dataset $D$ on which hypothesis search and verification is performed as a collection of tuples $\{\mathbf{x}_i\}_{i=1}^m$ that 
supports
multiple hypothesis semantic trees
resulting in a \textit{semantic forest} $\mathcal{F}:=\cup_i\mathcal{T}_{h^{(i)}}$, where each $\mathbf{x}_i$ is a row in the dataset and $x \in \mathbf{x}_i$ is an observation for a particular column. Further, $\mathbf{x}_i$ may span only a subset of nodes in $\mathcal{F}$, i.e., not all nodes in $\mathcal{F}$ may be observed. 
Specifically, while roots and leaves are always observed, internal nodes (target variables for intermediate hypotheses) may be latent.
Therefore, multiple versions of $D$ may be collected for $\mathcal{F}$ with different degrees of observability of internal nodes, altering the difficulty of the discovery task.

%% file: sections/3_discovery_bench.tex
\newcommand{\hmsscore}{\mathrm{HMS}}
\newcommand{\contextscore}{\mathrm{ctx}_\mathrm{F1}}
\newcommand{\variablescore}{\mathrm{var}_\mathrm{F1}}
\newcommand{\relationscore}{\mathrm{rel}_\mathrm{acc}}

\section{\discoverybench}
\label{sec:discovery_bench}

We now introduce a novel benchmark, \discoverybench{}, for discovering data-driven hypotheses. In this benchmark, a \emph{data-driven discovery task} is defined as follows: Given one or more task dataset(s)
$D$ and a discovery goal $G$, derive a hypothesis $h = \psi(c,v,r)$ addressing $G$ with the highest specificity for the context $c$, variables $v$, and relationship $r$ supported by $D$. Optionally, a workflow of deriving such a hypothesis can be outputted to augment information already present in the hypothesis. \discoverybench{} has two components: \textbf{\real}
encompassing data-driven hypotheses and workflows derived from published scientific papers and \textbf{\synth} capturing systemic variations in data-driven hypotheses and workflows obtained from synthetically generated datasets. We release our dataset under the ODC-BY license: \url{https://github.com/allenai/discoverybench}.


\subsection{\real: Collecting data-driven hypotheses in the wild}

Our goal is to replicate the scientific process undertaken by researchers to search for and validate a hypothesis from one or more datasets. We focus on six scientific domains where data-driven research is the cornerstone of scientific progress: sociology, biology, humanities, economics, engineering, and meta-science. Our data collection follows either a \textbf{data-first} or \textbf{code-first} approach.

For the \textbf{data-first approach}: 1) we filter papers based on open public datasets ($D$) such as National Longitudinal Surveys (NLS), Global Biodiversity Information Facility (GBIF), and World Bank Open Data (WBOD) that have workflow details; 2) we then try to replicate these workflows in Python. For this data-first approach, replication took up to 90 person-hours per dataset, often (30\%) not resulting in success. This highlights building data-driven discovery benchmarks from real studies is not only challenging and time-consuming, but automating discovery can also be key for scientific progress and reproducibility. 


The data-first approach by design is limited to well-known aforementioned public datasets. To improve diversity in domains, datasets ($D$), and workflows, we also adopted a \textbf{code-first approach} to look beyond popular public datasets. 
In this approach, we 1) search for code repositories based on scientific papers with available datasets and 2) attempt to replicate them in Python with existing code or from scratch with interpretation of the associated paper. We looked at 785 data points in Zenodo, EU's Open Research Repository, with a filter for computational notebooks. Over 85\% of the repositories either had missing code, code that could not be easily translated to Python, or a proprietary/non-open dataset. We finalized a candidate list of 14 repositories, but in the end, 3 of them passed all our checks for their hypotheses to be included in the benchmark\footnote{Some repositories include hypotheses from multiple papers as their background.}. 

\begin{figure}[t!]
    \centering
    \includegraphics[trim=0 0 0 0,clip, width=0.95\textwidth]{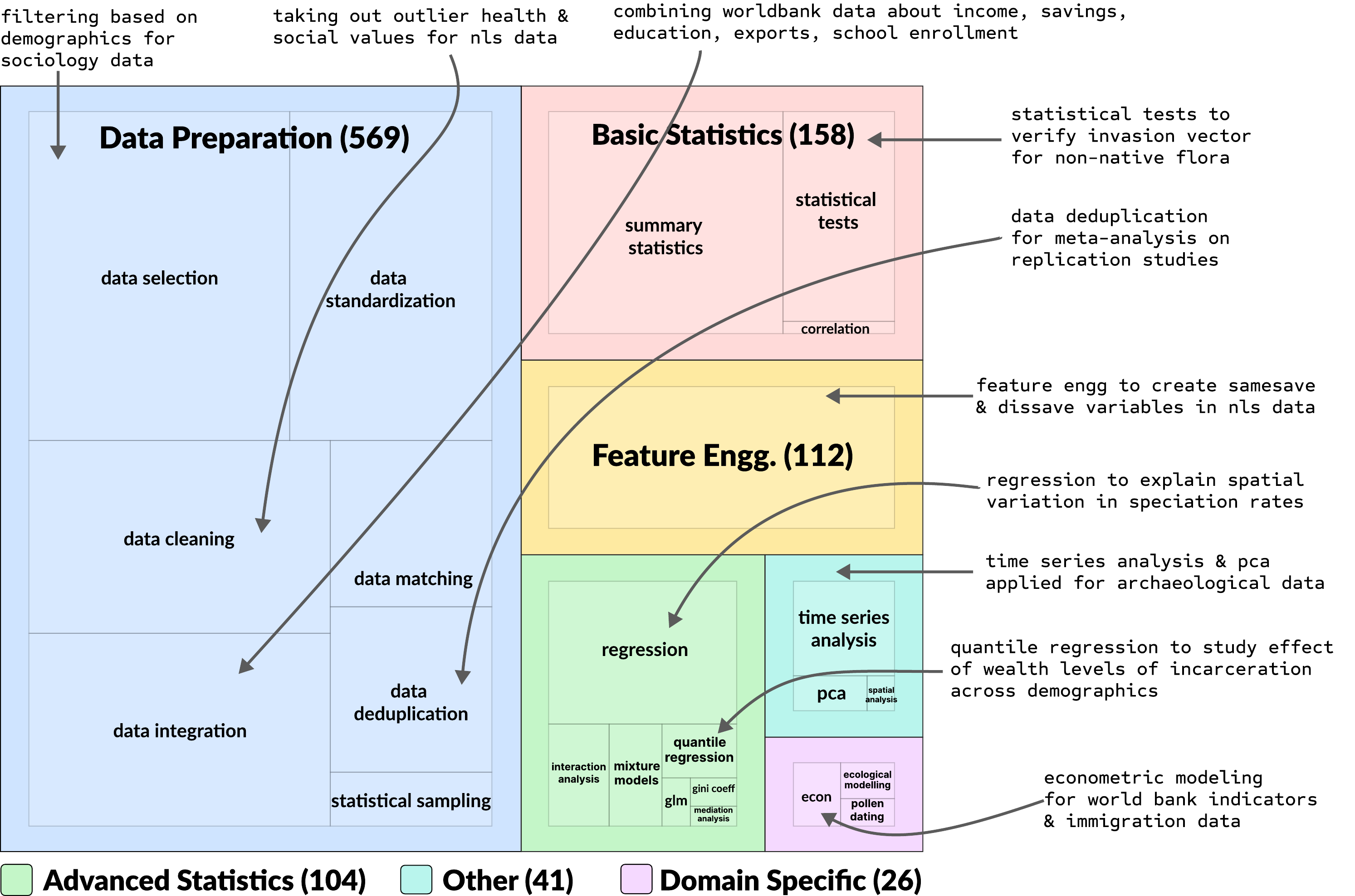}
    \caption{\small Workflow categories in \real{} with representative examples.}
    \label{fig:fig2}
    \vspace{-3mm}
\end{figure}



Upon replication of the result or implementation of the full procedure as described in the paper, we include the (dataset $D$, hypothesis $h$, implementation workflow) tuple to the benchmark.

During the process, the implementation workflow may lead to other hypotheses that are not directly reported in the paper but can be supported by the data.
We included them in \discoverybench{}, which leads to a good mix of already reported science-worthy hypotheses as well as novel
hypotheses grounded in datasets. This is particularly useful as our goal is to evaluate LLMs' ability to solve a discovery task that is realistic but never reported before.

Finally, the task datasets are supplemented with a dataset description, natural language descriptions of the columns, and additional background knowledge related to the domain or the datasets. Some of our tasks, for instance, archaeology, require domain knowledge to derive a particular hypothesis.





\newpara{Inferring task difficulty.}
Using the hypothesis semantic tree defined in Section~\ref{sec:background}, we say that the difficulty of a discovery task is proportional to the \textit{path length} from an observed node to the target hypothesis node in the tree.
However, knowing the tree structure from a task dataset alone is impractical due to incomplete \textit{a priori} information about unobserved intermediate nodes and edges between observed nodes. 
To infer task difficulty, we, therefore, approximate the path length between the target and leaf nodes using the length of the implementation workflow required to derive a target hypothesis. Specifically, for each step in the workflow, we add $1$ to the discovery path length.
In some cases, we derive two tasks: easy and hard from the same hypothesis, where for easy, we provide the derived variables as observed variables in the dataset (e.g., BMI), and for hard, it would require deriving intermediate variables (BMI from height and weight) to reach the target.
Additionally, given the view of a task dataset as encoding the union of multiple semantic trees rooted at different hypotheses, i.e., a semantic forest $\mathcal{F}$, we further posit that task difficulty increases as the number of trees in the forest ($|\mathcal{F}|$) increases. Intuitively, discovery becomes harder as the hypothesis search space increases. In practice, this setting is observed when a task requires access to multiple datasets. 









\newpara{Forming discovery goals.}
By definition, each hypothesis can be fully specified by the declarative sentence as $h:= \psi(c, v, r)$. To systematically construct the discovery goals for the task, we first mask one of each dimension, context $c$, variable $v$, relationship $r$, and generate a discovery goal to identify the masked information given the rest of the hypothesis and the task dataset(s). For instance, for a target hypothesis, \emph{``The effect of socioeconomic status on college degree completion is higher for females (0.4995) than males (0.4467)''}, we form a discovery goal as \emph{``How does socioeconomic status impact on college degree completion for females compared to males?''} seeking the relationship $r$ to be discovered from the dataset(s) given the relevant variables $v$ and context $c$. Additionally, we ensure each discovery goal leads to only one answer, i.e., the target hypothesis. 




\subsubsection{Features of \real{} benchmark}

\begin{wraptable}[8]{r}{0.4\textwidth}
\vspace{-1.2em}
\centering
\small
\resizebox{\linewidth}{!}{%
\begin{tabular}{@{}lcc}
\toprule
 & \bf Train & \bf Test \\
\midrule
\# tasks & 25 & 239 \\
\# unique hypotheses & 14 & 144 \\
\# tasks need $>1$ dataset & 4 & 110 \\
\# domains &  3 & 6 \\
\bottomrule
\end{tabular}%
}\vspace{-2pt}
\caption{\small Statistics for \real{}} \label{tab:real-stat}  \vspace{-3pt}
\end{wraptable}
\discoverybench{} incorporates a broad landscape of data-driven discovery.
With over 500 instances of data preparation activities such as cleaning, deduplication, and integration, captures the complexity of real-world data preprocessing for discovery.
Tasks also demand a spectrum of statistical methods, from statistical tests to mixture models, and include domain-specific approaches in econometric and ecological modeling, as reflected in the Fig~\ref{fig:fig2}\footnote{A task may require multiple data preparation and analytical activities}.





Table~\ref{tab:real-stat} shows the diversity of tasks both in train and test split for \real{}. Most importantly, the benchmark incorporates 114 (4 + 110) tasks that require more than one related datasets
to be analyzed, with a maximum of 6 datasets 
for a task. Each workflow within the dataset can be viewed as a composition of unit actions---such as code generation for statistical tests---that LLMs excel at, showing how our tasks require the chaining of such atomic actions to address complex scenarios for data-driven discovery. We measure the complexity of these workflows by quantifying the number of unit actions involved, referring to this as the \emph{workflow length}, whose distribution can be seen in Fig~\ref{fig:analysis}.









\subsection{\synth: Generating data-driven hypotheses using LLMs}
To scale data collection, we next introduce a supplementary benchmark,
which is synthetically constructed to enable controlled model evaluations. 
Our goal is to reverse-engineer the process of hypothesis discovery to synthesize datasets and discovery tasks of varying difficulty that require analysis workflows similar to those in the real-world benchmark. 
Our approach leverages the broad pre-trained knowledge of LLMs in four stages:

\textbf{Domain sampling:} First, we prompt the model to generate a list of diverse topics or \emph{domains} along with their natural language descriptions. E.g., 
``Ancient architecture'' $\rightarrow$ ``Related to historic buildings, architectural marvels, and ancient construction techniques''.

\textbf{Semantic tree construction:} For each domain, we then build a semantic tree $\mathcal{T}_h$,
recursively deriving nodes starting from a primary hypothesis $h$. Specifically, we prompt the model with the domain and a sampled real-world workflow (e.g., ``within-cluster analysis'') to generate a hypothesis and its target variable.
Setting the target variable as root, we then derive child nodes by generating the independent variables required to verify $h$ using $\mathcal{V}(\cdot)$.
We operationalize this by generating a column name and description for each child node (along with a data type and range) and a pandas expression\footnote{The pandas expression encodes the structured hypothesis $\psi(c,v,r)$.} \citep{mckinney-proc-scipy-2010} over only independent variables in $\mathcal{T}_h$ such that its execution results in the target variable. We repeat this with each leaf in $\mathcal{T}_h$ as the root of a new semantic sub-tree, generating intermediate hypotheses and a new set of variables until the desired height of $\mathcal{T}$ is reached.\footnote{In practice, with probability 0.6, we choose whether a node is derived further or marked as a leaf.}
We also generate a set of distractor columns disjoint from nodes in $\mathcal{T}$, thus resulting in a synthetic semantic forest $\mathcal{F}$.

\textbf{Data generation:} We then construct a task dataset $D:=\{\mathbf{x}_i\}_{i=1}^m$ by generating synthetic data in a bottom-up manner (i.e., from leaves to root) for each node in $\mathcal{F}$. Starting with various sampling strategies for leaf nodes (see more in Sec~\ref{sec:synth}),
for each subsequent level in $\mathcal{F}$, we create new columns for nodes by simply executing their pandas expressions.
Finally, to mimic real-world challenges in data collection, we probabilistically perturb each instance $x \in \mathbf{x}_i$ 
by adding noise or dropping values 
to create missing data\footnote{Each value is noised independently; therefore, each row has sufficient true data useful for discovery.}. 
Note that at this stage, $D$ contains a column for each node in $\mathcal{F}$.

\textbf{Task generation:} For each internal node $h$ in $\mathcal{F}$, we now create multiple task datasets $D_h^{(l)}$ from $D$, varying the difficulty of the discovery task based on the path length $l$ between $h$ and the observed independent variables in $\mathcal{F}$. 
Finally, we follow the same strategy for goal formulation as \real{}.
We generate 903 tasks over 48 diverse domains and assign them to train, dev, and test sets using a 60/20/20 split, where each task is additionally tagged with a difficulty level from 1-4. While we evaluate our agents on the test, the training set can serve as supervised data for improving models.

\subsection{Evaluation}
We evaluate task performance by measuring the alignment of the predicted and gold hypotheses in natural language.\footnote{We deliberately take an outcome-based approach as $> 1$ discovery path may lead to the same hypothesis.} We take inspiration from recent works in LLM benchmarking \cite{Shashidhar2023DemocratizingLA,Zeng2023EvaluatingLL,Yuan2023EvaluatingIL,Fu2023GPTScoreEA,alpaca_eval,wildbench2024} and design a model-based evaluation strategy using \texttt{gpt-4-preview-0125} as the \emph{evaluator}, conditioned on our structured formalism of data-driven hypotheses.


Recall the propositional form $h := \phi(h_1,\ldots,h_n)$ of a hypothesis $h$ that decomposes it into sub-hypotheses. We first use our GPT-4 based evaluator to independently decompose the gold ($h^g$) and predicted ($h^p$) hypotheses into their respective sub-hypotheses $\{h^g_i\}_{i=1}^n$ and $\{h^p_j\}_{j=1}^m$,
asking it to also identify, for each sub-hypothesis $h_k$,
its context, variables, and relationship dimensions (prompt in \Cref{list:decompose}). Given this structured representation of the gold and predicted hypotheses, we then compute a \textbf{hypothesis match score (HMS)}, which measures the degree to which two hypotheses align on each dimension, as follows.

To compute HMS, we match each predicted sub-hypothesis $h^p_j$ with a gold sub-hypothesis $h^g_i$ when their contexts are judged as equivalent by our GPT-4 based evaluator (prompt in \Cref{list:match}).\footnote{Note that at most one gold sub-hypothesis is matched with a predicted sub-hypothesis.} Let $M$ denote this set of context-matched pairs of predicted and gold sub-hypotheses. At this point, treating each sub-hypothesis context as a single unit, we can compute an F1 score, $\contextscore$, capturing how aligned the $n$ contexts of sub-hypothesis of $h^g$ with the $m$ contexts of sub-hypotheses of $h^p$. Then, for each matched pair of sub-hypotheses, we measure how well the variables and relations align, using an F1 score for the variables ($\variablescore$) and an accuracy score for the relation ($\relationscore$).
Specifically, for each sub-hypothesis pair in $M$, we extract the set of interacting variables
in the gold and predicted sub-hypotheses using the GPT-4 based evaluator (prompt in \Cref{list:variable}). We compute the alignment between these two sets of variables as an F1 score, $\variablescore$, similar to how $\contextscore$ was computed. For relationships, we compute relationship accuracy with reference to the relationship between the gold variables
($\relationscore$) based on evaluator judgments using the following scoring heuristic: $100$ if there is an exact match of the relation, $50$ when the predicted relationship is broader than the gold relationship but encompasses it, and $0$ otherwise (prompt in \Cref{list:rel}).
Finally, we compute HMS $\in [0, 100]$ as the average alignment of the variable and relationship dimensions over context-matched sub-hypotheses, weighted by the overall context alignment:
\[
    \hmsscore(h^p, h^g) = \contextscore(h^p, h^g) \times \frac{1}{|M|} \sum_{i=1}^{|M|} \Big( \variablescore(h^p_i, h^g_i) \times \relationscore(h^p_i, h^g_i) \Big)
\]

%% file: sections/5_experiments.tex
\begin{figure*}[t]
\begin{minipage}[h]{.5\textwidth}
\small
\centering
\resizebox{0.9\linewidth}{!}{%
\begin{tabular}{l|c|c|c}
\toprule
            & \bf {GPT-4o} & \bf {GPT-4p} & \bf {Llama-3} \\
\midrule
\multicolumn{4}{l}{\bf \real{}}   \\
\midrule
NoDataGuess & 0.0  &      4.7 &   11.5 \\
CodeGen     & 15.5  &      16.3 &   12.1 \\
React       & 15.4  &      15.6  &  13.5 \\
DataVoyager & 15.4  &      13.9  &  11.5   \\
Reflexion (Oracle)   & \bf 24.5  &  \bf 19.5  & \bf 22.5 \\

\midrule
\multicolumn{4}{l}{\bf \synth{}}                                                                                \\
\midrule
CodeGen       & 14.1 &        8.7 &       10.9 \\
React         & 11.6 &        7.4 &       12.0 \\
DataVoyager   & 5.7  &        6.9 &        11.7 \\
Reflexion (Oracle)  & \bf 15.7 &   \bf  12.9 &  \bf  23.2  \\
\bottomrule
\end{tabular}%
}
\end{minipage}%
\hspace{0em}
\begin{minipage}[h]{.5\textwidth}
 \centering
\centering
  \includegraphics[trim=0 0 0 0,clip, width=\textwidth]{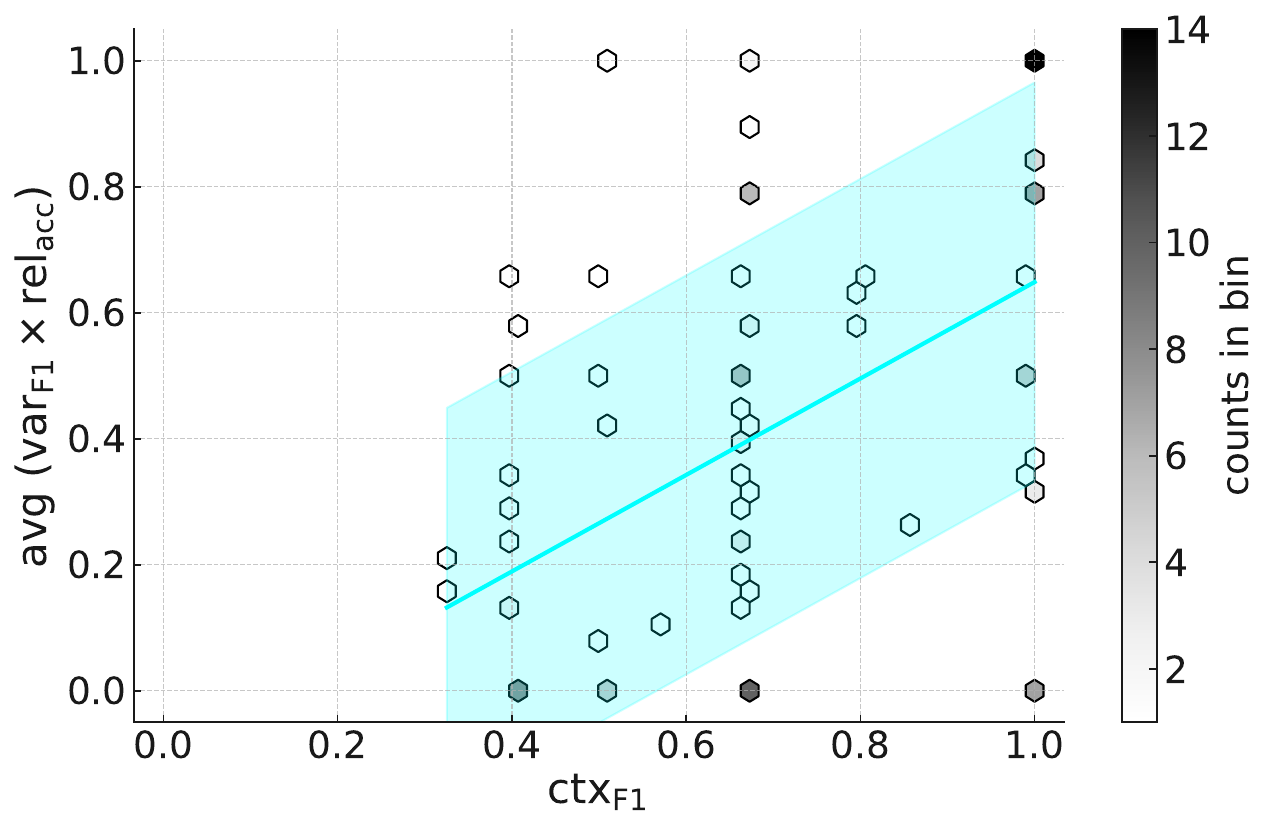}\\
\end{minipage}
 \caption{\small (Left) Hypothesis Matching Scores ($\mathrm{HMS}$) across agent-LLM pairs in \real{} and \synth{}. (Right) Scatter plot for $\contextscore$ and average $\variablescore \times \relationscore$, showing accurate contexts increases the probability of predicting variables and relations accurately. Scores are for the best model on \real{} and only include data points (44.2\%) where both scores are non-zero.}
    \label{fig:main-results}
    \vspace{-2mm}
\end{figure*}

\section{Experiments}
\label{sec:experiments}

\subsection{Discovery Agents}
\label{sec:discovery_agents}
We benchmark state-of-the-art LLM-based few-shot reasoning methods as discovery agents with two closed models, GPT-4o and GPT-4-0125-preview (GPT-4p), and one open, Llama-3-70B, model powering the reasoning methods. A discovery agent takes the task description, paths to the task dataset(s) $D$, metadata about the datasets (description, column descriptions), and the goal, $G$, to produce a natural language (NL) hypothesis specified by context, variables, and relationship. 


\begin{ite}
    \item \textbf{CodeGen} generates the entire code at one go to solve the task, where we provide a demonstration of a solution code in the context. After code execution and based on the result, it generates the NL hypothesis and summarizes the workflow.
    \item \textbf{ReAct} \cite{yao2023react} solves the task by generating thought and subsequent codes in a multi-turn fashion.
    \item \textbf{DataVoyager} is a multi-component data-driven discovery agent from \cite{majumder2024data}. It has four components, planner, code generator, data analysis, and critic, that orchestrate the discovery process.
    \item \textbf{Reflexion (Oracle)} \cite{Shinn2023ReflexionLA} is an extension of CodeGen agent, where at the end of one trial, we provide the ``oracle'' $\hmsscore$ score as an evaluation signal, and it generates a reflection to improve (when $\mathrm{HMS} < 1$) in the next trial till it solves the task, or maximum trials (3) are reached.
    \item \textbf{NoDataGuess} guesses the hypothesis (in \real{}) just from the dataset description and the goal without accessing the datasets where we measure LLM's memorization of already published works.
\end{ite}

\subsection{Main Results}
Fig~\ref{fig:main-results}(left) shows that overall performance for all framework-LLM pairs is low for both \real{} and \synth{}, highlighting the challenging nature of the task and the benchmark. Most importantly, effective reasoning prompts such as React and planning with a self-critic (DataVoyager) do not help improve the simple CodeGen agent. But with oracle feedback, Reflexion (Oracle) significantly improves over CodeGen (base) performance.  
Analysis reveals that almost all non-reflexion agents solve the \emph{easiest} (in terms of workflow category and length) instances from the benchmark. GPT-4o refuses to hallucinate in the NoDataGuess baseline, whereas surprisingly Llama-3 performs similarly in both data and no-data modes. 
We additionally observe that the models' performance in \real{} and \synth{} are similar, indicating our synthetic benchmark captures complexities of the real workflow but provides a systematic way to analyze the models' performance.

\subsection{Analysis}

\textbf{Context is important.} Fig~\ref{fig:main-results}(right) shows the trends of the $\contextscore$ and combined $\variablescore \times \relationscore$. A positive trend signifies that to predict variables and relationships accurately, precise and accurate context prediction is necessary. However, correct identification of context is an important first step, although it does not guarantee success. 

\textbf{Workflow complexity barrier.} Almost all agents struggle more with tasks involving complex statistical techniques, complex data preparation methods, or domain-specific models. The top three workflow categories where the best non-oracle model was highly performant are correlation analysis (55\%), data selection (18\%), and summary statistics (18\%), whereas the lowest three workflow categories are spatial analysis (0\%), pollen dating (0\%), and ecological modeling (0\%).

\begin{figure}[t!]
    \centering
    \includegraphics[trim=15 260 10 260,clip, width=\textwidth]{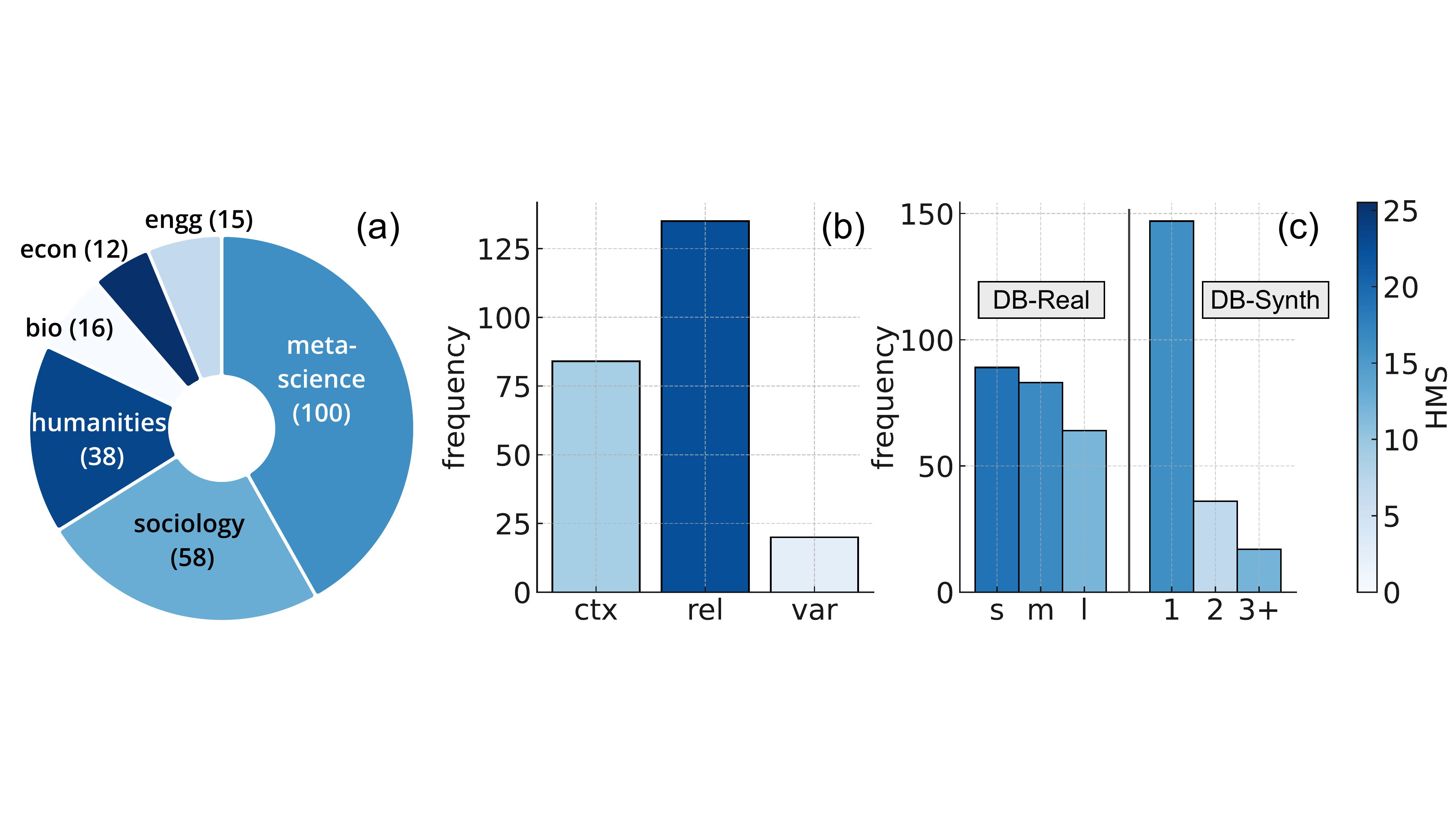}
    \caption{Best non-oracle agent's performance ($\hmsscore$) (a) across domains, (b) for goal types (dimension to be discovered), and (c) for different workflow lengths. In (c) workflow length categories for \real{} are s: $<10$, m: $>10, <20$, l: $>20$. For \synth{}, it is the semantic tree height.}
    \label{fig:analysis}
\end{figure}

    
\textbf{Domain knowledge dependency.} To check if additional domain helps agents perform better, we collect targeted domain knowledge for the archaeology-related tasks that needed significant domain knowledge during data collection. When added as additional hints, we find that DataVoyager's (GPT-4p) performance jumps from 9.9\% (w/ domain knowledge) to 17.5\% (w/o domain knowledge).


    
    
    

\textbf{Performance across domains and goal types.} Fig~\ref{fig:analysis}(a) depicts that biology (0\%) and engineering (7\%) perform the worst due to their higher dependence on advanced statistical methods, while economics (25\%) and sociology (23\%) perform better. Additionally, Fig~\ref{fig:analysis}(b) shows goals related to discovering a relationship given context and variables are more easily solved than the other two types of goals, finding context and variables. This is explained by the complexity of the hypothesis search, which is broader for finding the right context or a set of variables given a fixed relationship, whereas finding the relationship given context and variables is easier.

\textbf{Impact of workflow length.}
Inherently, the difficulty of the tasks is measured by the gold workflow length (\real{}) or the height of the semantic tree (\synth{}). \Cref{fig:analysis}(c) shows a decreasing trend in performance as workflow length (hence, complexity) increases. The performance drops significantly even for medium-length workflows, highlighting current agents' limitations.  



%% file: sections/7_conclusion.tex
\section{Conclusion}
\label{sec:conclusion}
We present \discoverybench, the first data-driven discovery benchmark consisting of 264 discovery tasks that capture real scientific workflows extracted from published works. We supplement this with 903 structurally generated synthetic tasks, tailored to evaluate discovery agents at various levels of difficulty. We benchmark state-of-the-art reasoning frameworks with the most advanced LLMs, but the best agent's performance only peaks at 25\% underscoring the challenging nature of the task and the benchmark. We hope our timely contribution can increase interest and efforts in making progress on reliable and reproducible autonomous scientific discovery using large generative models. 



%% file: sections/8_appendix.tex
\appendix

\section{FAQs}
\begin{enumerate}
    \item \textbf{Dataset or Benchmark:} Is this a dataset or a benchmark? {\textbf{A benchmark}}
    
    \item \textbf{Benchmark:} For benchmarks, the supplementary materials must ensure that all results are easily reproducible (i.e., all necessary datasets, code, and evaluation procedures must be accessible and documented)
    
    \textbf{Datasets:} \discoverybench{} is released at \url{https://github.com/allenai/discoverybench/tree/main/discoverybench} . 

    \textbf{Code (Baseline Models):} Code for Discovery Agents are provided in the repository, at: \url{https://github.com/allenai/discoverybench/tree/main/agents}. A CLI is available to run the discovery agents on the benchmark. 

    \textbf{Evaluation Procedures:} Please follow our main paper for the details of our evaluation process. The code to run eval on a single instance of our benchmark is provided at: \url{https://github.com/allenai/discoverybench/tree/main/eval}. A CLI and some example scripts have been provided as well. 

    \item{\textbf{Accessibility}: The following are accessibility items on the submission checklist:}

    \textbf{Links to access the benchmark:} The link to access the benchmark is provided in the main submission (\url{https://github.com/allenai/discoverybench/tree/main/discoverybench}). 

    \textbf{Any data should use open and widely used formats. Simulation environments should explain how they can be used:} Our data are stored in widely accessible standard formats (e.g., \textsc{JSON, CSV}), with the structure described in  \Cref{sec:datadetails}.  

    \textbf{Long-term preservation.} Code and data are provided on \textsc{Github}.  All aspects will be publicly available for a long term.

    \textbf{Explicit Licence:} Our benchmark is licensed using \textsc{ODC-BY} and the associated code is licensed with \textsc{Apache 2.0}, as included in the \textsc{Github} repository. 

    \textbf{Structured Metadata for a dataset:} Our dataset is also available as the HuggingFace dataset: \url{https://huggingface.co/datasets/allenai/discoverybench}. Structured Metadata will be available once we finalize our work after addressing the reviewers' comments, if any.

    \textbf{A persistent dereferenceable identifier (e.g., a code repository such as GitHub):} The repository for our benchmark is: \url{https://github.com/allenai/discoverybench}.

\end{enumerate}

\section{Limitations}
\label{sec:limitations}
We currently filtered domains and tasks that required forecasting, simulation, or very specific modeling (species distribution, infection spread, astrophysics equations for exoplanets) in the benchmark as they were very time-consuming to replicate as well as discover hypotheses. As a result, we discarded more papers focused on natural and physical sciences compared to social sciences, which we plan to include in future benchmarks. 

We currently do not tackle the challenge of understanding and processing massive datasets, such as the 8.92 petabytes data from the Cancer Genome Atlas (\url{https://portal.gdc.cancer.gov}) or the extensive brain data from the Allen Institute (\url{https://alleninstitute.org/division/brain-science}). While the potential to discover new insights from such vast data volumes is significant, ensuring these findings are robust and not subject to $p$-hacking remains unaddressed by our current methods.

We currently do not handle multi-modal data and complex pipelines, such as those needed for analyzing satellite and other geospatial data relevant to climate science and astronomy data. This would involve multiple stages of data processing, the use of various tools, and managing workflow complexities, for example, analyzing thousands of species patterns combined with satellite data to study habitats. So we do not incorporate workflows like those of EarthRanger (\url{https://www.earthranger.com}).

\newpara{Ethical Considerations} There could be many potential societal consequences of systems tuned on our proposed benchmark since it involves using LLMs, such as policy misuse, legal ramifications, and false discovery. On the positive side, our proposed benchmark can advance the rate of discovery, leading to an improved standard of living and social well-being.

\section{Data collection for \real{}}
\label{sec:real}
For data-first approach, replication took 15 to 40 person-hours for each NLS-related paper and up to 90 person-hours
for the GBIF dataset, where specialized domain knowledge and tools led to higher complexity. All papers replicated in the NLS dataset were included, while less than half of the papers in specialized datasets like GBIF and WBOD were added to \discoverybench{}. 

\textbf{Citation/Repositories for \real{}:}
List of scientific works from where we have replicated our gold workflows and hypotheses:
\begin{enumerate}
    \item Sociology: \cite{zaw2016race,apel2010impact,alexander1982social, smith2005time,dougherty2003numeracy}
    \item Biology: \cite{cerezer2023accelerated, riera2024effect}
    \item Economics: \cite{pal2023impact,appiah2017effect,weatherly2022impact,rambeli2021dynamic,ottaviano2013immigration}
    \item Engineering: \cite{alves2023status}
    \item Meta-science: \cite{heyard2024meta}
    \item Humanities: \cite{brozio2024patterns,brozio2019monuments,lorenz2018kommunikationsstrukturen,maida2023unter,sommerfeld2013gerategeld,feeser2019human,parkinson2021radiocarbon,kneisel2021chronology,kaiser2021time,brinkmann2019copper,dal2019s,palmisano2021long,parkinson2021radiocarbon}
\end{enumerate}
All assets come under CC license or open licenses.

\section{Data Generation for \synth{}}
\label{sec:synth}
For leaves, we use different sampling strategies based on the data type. 
Specifically, for categorical nodes, we sample instances with replacement from the range of allowed values, whereas for numeric, we first select a distribution (e.g., normal) and its parameters based on the specified range and then perform sampling. 
For each subsequent level in $\mathcal{F}$, we create new columns for nodes by simply executing their pandas expressions\footnote{The expression is guaranteed to only have variables already generated due to the bottom-up construction.}. 
To recover from any execution errors, we additionally use a self-refine \citep{madaan2023selfrefine} approach to generate new pandas expressions guided by the execution error logs. 
Finally, to mimic real-world challenges in data collection, we probabilistically perturb each instance $x \in \mathbf{x}_i$ 
by adding noise or dropping values 
to create missing data\footnote{Each value is noised independently; therefore, each row has sufficient true data useful for discovery.}. 
After generation, $D$ contains a column for each node in $\mathcal{F}$.

\section{Datasheets}
\label{sec:datasheets}

\subsection{Motivation}

\begin{itemize}
    \item \textbf{For what purpose was the dataset created?} \discoverybench{} is created to help assess large language models' (LLMs) ability to automate the search and verification of hypotheses purely from a set of provided datasets.
    \item \textbf{Who created the dataset (e.g., which team, research group) and on behalf of which entity (e.g., company, institution, organization)?} Authors belong to the Allen Institute for AI, OpenLocus, and the University of Massachusetts Amherst. The data collection is part of research efforts conducted by the Allen Institute for AI.
    \item \textbf{Who funded the creation of the dataset?} Allen Institute for AI.
\end{itemize}



\subsection{Collection Process}

\begin{itemize}
    \item \textbf{How was the data associated with each instance acquired?}
    Our goal is to replicate the scientific process undertaken by researchers to search for and validate a hypothesis from one or more datasets. We focus on six scientific domains where data-driven research is the cornerstone of scientific progress: sociology, biology, humanities, economics, engineering, and meta-science. Our data collection follows either a \textbf{data-first} or \textbf{code-first} approach. Each instance has been manually implemented and verified by the authors for solvability.
\end{itemize}

\subsection{Uses}
\begin{itemize}
    \item \textbf{Has the dataset been used for any tasks already?}
    We use this benchmark to evaluate LLM's ability to search and verify hypotheses purely from a set of datasets.
    \item \textbf{Are there tasks for which the dataset should not be used?} We do not expect the community members to use this data to train models that can aggravate $p$-hacking.
\end{itemize}

\subsection{Distribution and Maintainance}
\begin{itemize}
    \item \textbf{How will the dataset will be distributed?}
    We distribute this benchmark via our \textsc{Github} repository: \url{https://github.com/allenai/discoverybench} and \url{https://huggingface.co/datasets/allenai/discoverybench}.
    \item \textbf{How can the owner/curator/manager of the dataset be contacted?} For any benchmark-related queries, please contact: \texttt{bodhisattwam@allenai.org}. For any code-related discussions, please raise an issue in \textsc{Github}: \url{https://github.com/allenai/discoverybench}.
\end{itemize}

\section{Composition of \discoverybench{}}
\label{sec:datadetails}



\subsection{Metadata structure}
\begin{itemize}
    \item \textbf{id}: An identifier for the metadata.
    
    \item \textbf{domain}: The broad field of study or area of research.
    
    \item \textbf{workflow\_tags}: A set of keywords summarizing the main processes or techniques used in the replication implementation. They provide an overview of the methodological approach and facilitating the identification of relevant analytical techniques.
    
    \item \textbf{domain\_knowledge}: 
    \begin{itemize}
        \item Contextual information or insights related to the dataset, explaining how certain behaviors or variables can be interpreted within the field of study.
        \item It helps open avenues to think in directions that LLM might not have considered otherwise, broadening the understanding of the field.
    \end{itemize}
    
    \item \textbf{datasets}: Contains detailed information about the datasets used, including:
    \begin{itemize}
        \item \textbf{name}: The name or filename of the dataset.
        
        \item \textbf{description}: A summary of the dataset's contents and the type of data it includes.
        
        \item \textbf{max\_depth}: The maximum hierarchical level of nested data structures within the dataset, indicating the complexity of the data.
        
        \item \textbf{columns}: Detailed descriptions of each column in the dataset, including:
        \begin{itemize}
            \item \textbf{name}: The column's name or header.
            \item \textbf{description}: Explanation of the data contained in the column and its significance.
            \item \textbf{depth}: The hierarchical level of the column within the dataset, indicating its structural position.
        \end{itemize}
    \end{itemize}
    
    
    \item \textbf{hypotheses}: Statements or predictions being tested, divided into:
    \begin{itemize}
        \item \textbf{main}: Primary hypotheses that are central to the discovery task.
    \end{itemize}
    
    \item \textbf{workflow}: A step-by-step description of the replication process followed to validate the hypotheses, outlining the methods and procedures used from data preparation to final analysis. Some of the workflows and sub-workflows are high-level and thus the same for different queries as they follow the same implementation leading to a range of hypotheses. 
    
    \item \textbf{queries}: Goals related to each hypothesis, each including:
    \begin{itemize}
        \item \textbf{qid}: A unique identifier for the goal for a given true/gold hypothesis.
        \item \textbf{difficulty}: Categorization of the difficulty. Structurally defined for \synth{} using the semantic tree definition.
        \item \textbf{true\_hypothesis}: The hypothesis being tested through the goal. This defines the primary statement or prediction under investigation.
        \item \textbf{relevant\_cols}: Columns from the dataset that are relevant to answering the query, indicating the specific data points that can be used in the analysis. Only appears for \synth{}.
        \item \textbf{target\_col}: The column being predicted or the dependent variable in the analysis. Only appears for \synth{}.
        \item \textbf{question\_type}: The type of question being asked categorizing the nature of the inquiry.
        \item \textbf{question}: The discovery goal.
    \end{itemize}
\end{itemize}

\subsection{Directory structure for \real{}}


There may be more than one query per metadata. The train split contains 14 metadata files and 25 queries. The test split contains 144 metadata files and 239 queries. Metadata folders with the same prefixes use the same underlying dataset with either a subset or a preprocessed version. When dealing with a full dataset (i.e., nls\_raw), the task becomes substantially harder due to the data preparation required.

\begin{verbatim}
    |-test
    |---archaeology
    |---introduction_pathways_non-native_plants
    |---meta_regression
    |---meta_regression_raw
    |---nls_incarceration
    |---nls_raw
    |---nls_ses
    |---requirements_engineering_for_ML_enabled_systems
    |---worldbank_education_gdp
    |---worldbank_education_gdp_indicators
    |-train
    |---evolution_freshwater_fish
    |---immigration_offshoring_effect_on_employment
    |---nls_bmi
    |---nls_bmi_raw
\end{verbatim}

\subsection{Directory structure for \synth{}}


There is one query per metadata. The train split contains 551 metadata files (queries), the dev split contains 153 metadata files (queries), and the test split contains 200 metadata files (queries).

\begin{verbatim}
   |-test
   |---ancient-languages_*_*
   |---artificial-ecosystems_*_*
   |---astronomy_*_*
   |---board-games_*_*
   |---coding-competitions_*_*
   |---digital-artistry_*_*
   |---futuristic-technology_*_*
   |---impressionist-art_*_*
   |---machine-learning_*_*
   |---molecular-gastronomy_*_*
   |---neuroscience_*_*
   |---philosophical-debates_*_*
   |---robotics_*_*
   |-train
   |---adventure-travel_*_*
   |---ancient-architecture_*_*
   |---ancient-astronomy_*_*
   |---aviation_*_*
   |---biodiversity-conservation_*_*
   |---cryptic-puzzles_*_*
   |---cryptocurrency_*_*
   |---culinary-arts_*_*
   |---cybersecurity_*_*
   |---environmental-activism_*_*
   |---fashion-design_*_*
   |---fine-arts_*_*
   |---literary-classics_*_*
   |---marine-biology_*_*
   |---marine-conservation_*_*
   |---medieval-literature_*_*
   |---musical-therapy_*_*
   |---photography_*_*
   |---robotic-explorers_*_*
   |---solar-power_*_*
   |---space-tourism_*_*
   |---steampunk-culture_*_*
   |---theater-productions_*_*
   |---underwater-archaeology_*_*
   |---urban-gardening_*_*
   |---vintage-automobiles_*_*
   |---virtual-reality_*_*

\end{verbatim}

\section{Discovery Agent}


The command \texttt{discovery\_agent.py} is used with various options to customize its behavior for discovery tasks. Below are the options explained:

\begin{itemize}[leftmargin=*]
    \item \textbf{Usage:} \texttt{discovery\_agent.py [OPTIONS] QUERY} -- Executes the discovery agent with specified options.

    \item \textbf{Options:}
    \begin{itemize}
        \item \texttt{--agent\_type [coder|react]}: Specifies the type of agent to use for discovery. The default type is \texttt{coder}. Options include \texttt{coder} for code-related tasks and \texttt{react} for reactive tasks.
        
        \item \texttt{--model\_name TEXT}: Sets the model to be used. The default is \texttt{gpt-4o}. Available models include \texttt{gpt-4-turbo}, \texttt{llama-3-70b-chat}, \texttt{claude-3-opus}, and \texttt{gemini-pro}. An exhaustive list is available in \texttt{config/model\_config.json}.
        
        \item \texttt{--api\_config TEXT}: Path to the API configuration file. The default path is \texttt{config/api\_config.json}.
        
        \item \texttt{--log\_file TEXT}: Specifies the path to the log file where operations details are stored.
        
        \item \texttt{--metadata\_path TEXT}: Path to the metadata file. This option is required.
        
        \item \texttt{--metadata\_type [real|synth]}: Specifies the type of metadata, where \texttt{real} stands for actual metadata and \texttt{synth} for synthetic. This option is required.
        
        \item \texttt{--add\_domain\_knowledge}: Includes domain-specific knowledge in the query processing.
        
        \item \texttt{--add\_workflow\_tags}: Includes workflow tags in the query to enhance context.
        
        \item \texttt{--help}: Displays the help message and exits, showing all available command options.
    \end{itemize}
\end{itemize}

\section{Evaluation}

Explain about evaluation in a line and then explain the CLI usage here. 

The command \texttt{discovery\_eval.py} is used to evaluate the outputs generated by the discovery agent. Below are the detailed descriptions of the command options:

\begin{itemize}[leftmargin=*]
    \item \textbf{Usage:} \texttt{discovery\_eval.py [OPTIONS] QUERY} -- Executes the evaluation agent with specified options and a query.

    \item \textbf{Options:}
    \begin{itemize}
        \item \texttt{--gold\_hypo TEXT}: Specifies the gold standard hypothesis for comparison. This field is required.
        
        \item \texttt{--gold\_workflow TEXT}: Specifies the gold standard workflow to be used as a reference during evaluation.
        
        \item \texttt{--pred\_hypo TEXT}: Specifies the predicted hypothesis generated by the discovery agent. This field is required.
        
        \item \texttt{--pred\_workflow TEXT}: Specifies the predicted workflow generated by the discovery agent.
        
        \item \texttt{--metadata\_path TEXT}: Specifies the path to the metadata file that is utilized during evaluation. This field is required.
        
        \item \texttt{--metadata\_type [real|synth]}: Determines the type of metadata used in the evaluation, where \texttt{real} indicates actual metadata and \texttt{synth} indicates synthetic metadata. This field is required.
        
        \item \texttt{--eval\_output\_path TEXT}: Specifies where the evaluation results should be saved.
        
        \item \texttt{--help}: Displays the help message and exits, detailing all available command options.
    \end{itemize}
\end{itemize}

\section{Experiments}
\label{app:exp}
For GPT-based models, we use OpenAI API (\url{https://platform.openai.com/docs/models}), and for Llama3, we used Together API (\url{https://docs.together.ai/docs/inference-models})

\section{Evaluator Prompts}
\label{app:evaluation-prompts}

We provide below the exact prompts used for our GPT-4 based evaluation of the generated hypothesis against the gold hypothesis.

\begin{listing}[!ht]
\begin{minted}[frame=lines,
baselinestretch=1,
bgcolor=Box3Color,
fontsize=\footnotesize, breaklines, breaksymbolleft={}, breaksymbolright={}]{python}
decomposition_prompt = f"""\
        Given a set of dataset columns, a ground-truth hypothesis, and the analysis workflow used, your task is to extract the set of sub-hypotheses that are present in the hypothesis such that each sub-hypothesis covers a separate context, is self-sufficient, and operates on a coherent set of 3 dimensions: Context, Variables, and Relations.
        
        Here are the definitions for these dimensions:
        
        - Contexts: Boundary conditions that limit the scope of a sub-hypothesis. E.g., “for men over the age of 30”, “in Asia and Europe”, or "None" if there is no boundary condition specified.
        
        - Variables: Known concepts that interact in a meaningful way under a given context to produce the sub-hypothesis. E.g., gender, age, income, or "None" if there is no interacting variable.
        
        - Relations: Interactions between a given set of variables under a given context to produce the sub-hypothesis. E.g., “quadratic relationship”, “inversely proportional”, piecewise conditionals, or "None" if there is no interacting relationship. 
        
        Make sure to only use the information present in the hypothesis and the workflow. Do not add any new information.
        If no sub-hypotheses can be extracted, return an empty list.

        Here is the metadata for the task:
        ```json
        {{
            "datasets": {dataset_metadata},
            "hypothesis": "{hypothesis}",
            "workflow": "{workflow}"
        }}
        ```

        Return your answer as a JSON object in the following format:
        ```json
        {{
        "sub_hypo": [
            {{
                "text": the sub-hypothesis in natural language,
                "context": a short text description of the context of the sub-hypothesis,
                "variables": a list of columns involved in the sub-hypothesis,
                "relations": a short text description of the relationship between the variables of the sub-hypothesis,
                "explanation": a short text explanation for the breakdown of the sub-hypothesis
            }},
            ...
        ]
        }}```
        """
\end{minted}
\vspace{-1em}
\caption{Decomposition Prompt to obtain sub-hypotheses from a hypothesis.}
\label{list:decompose}
\end{listing}

\begin{listing}[!ht]
\begin{minted}[frame=lines,
baselinestretch=1,
bgcolor=Box3Color,
fontsize=\footnotesize, breaklines, breaksymbolleft={}, breaksymbolright={}]{python}
matching_prompt = f"""
        Given a gold hypothesis, a gold context, a predicted hypothesis, and a predicted context, your task is
        to determine if the predicted context semantically matches the ground-truth context.
        
        Here is the definition for Context: Boundary conditions that limit the scope of a sub-hypothesis. E.g., “for men over the age of 30”, “in Asia and Europe”, or "None" if there is no boundary condition specified.
        
        If the predicted context matches the gold context, return true, otherwise return false.

        Here is the metadata for the task:
        ```json
        {{
            "gold_hypothesis": "{gold_hypotheis}",
            "gold_context": "{gold_context}",
            "predicted_hypothesis": "{pred_hypothesis}",
            "predicted_context": "{pred_context}"
        }}
        ```

        Return your answer as a JSON object in the following format:
        ```json
        {{
            "match": true or false
        }}
        ```"""
\end{minted}
\vspace{-1em}
\caption{Matching prompt to match contexts of two sub-hypotheses.}
\label{list:match}
\end{listing}

\begin{listing}[!ht]
\begin{minted}[frame=lines,
baselinestretch=1,
bgcolor=Box3Color,
fontsize=\footnotesize, breaklines, breaksymbolleft={}, breaksymbolright={}]{python}
main_context = f"""
        You are going to compare two natural-language hypotheses HypoA and HypoB accompanied with optional workflows: WorkflowA for HypoA and WorkflowB for HypoB.
        Both the hypotheses answer the natural language query "QUERY" over the dataset(s) described by dataset description(s) and column description(s) below.
        Compare HypoA and HypoB in terms of three aspects: Contexts, Variables, and Relations.
        E.g., for the hypothesis "From 1995 to 2009, the number of sandhill cranes around the tundra (Indigilka River) surged by an astounding ~10X":
        * Contexts refer to the stratification of the data under which the given hypothesis is True. E.g., "For all women", "From 1995 to 2009". 
        * Variables refer to the set of variables (either dependent or independent) that are mentioned in the hypothesis. E.g., number of sandhill cranes, location.
        * Relations refer to the form of relation between the variables. E.g., "surged by ~10x".

        Answer the following questions for a given pair of hypotheses, HypoA and HypoB, along with an explanation grounded on the QUERY and the DATASET(S).

        Here is the metadata for the task:
        ```json
        {{
        "datasets": {datasets_json},
        "query": {query},
        "HypoA": {gold_hypo},
        "WorkflowA": {gold_workflow},
        "HypoB": {gen_hypo},
        "WorkflowB": {gen_workflow}
        }}
        ```

        {variable_question}"""
variable_question = """\
        Question: For both HypoA and HypoB, what are the different variables found in the hypotheses? \
        Return your answer as a JSON object in the following format:
        ```json
        {{
        "sizeA": num of variables used in HypoA
        "sizeB": num of variables used in HypoB
        "intersection": num of variables common in HypoA and HypoB. Use *fuzzy matching* to determine intersection, accounting for paraphrases or slightly different surface forms
        "explanation": a short text explanation about the variables
        }}```
        Answer:"""
\end{minted}
\vspace{-1em}
\caption{Prompt for variable alignment between two sub-hypotheses.}
\label{list:variable}
\end{listing}

\begin{listing}[!ht]
\begin{minted}[frame=lines,
baselinestretch=1,
bgcolor=Box3Color,
fontsize=\footnotesize, breaklines, breaksymbolleft={}, breaksymbolright={}]{python}
main_context = f"""
        You are going to compare two natural-language hypotheses HypoA and HypoB accompanied with optional workflows: WorkflowA for HypoA and WorkflowB for HypoB.
        Both the hypotheses answer the natural language query "QUERY" over the dataset(s) described by dataset description(s) and column description(s) below.
        Compare HypoA and HypoB in terms of three aspects: Contexts, Variables, and Relations.
        E.g., for the hypothesis "From 1995 to 2009, the number of sandhill cranes around the tundra (Indigilka River) surged by an astounding ~10X":
        * Contexts refer to the stratification of the data under which the given hypothesis is True. E.g., "For all women", "From 1995 to 2009". 
        * Variables refer to the set of variables (either dependent or independent) that are mentioned in the hypothesis. E.g., number of sandhill cranes, location.
        * Relations refer to the form of relation between the variables. E.g., "surged by ~10x".

        Answer the following questions for a given pair of hypotheses, HypoA and HypoB, along with an explanation grounded on the QUERY and the DATASET(S).

        Here is the metadata for the task:
        ```json
        {{
        "datasets": {datasets_json},
        "query": {query},
        "HypoA": {gold_hypo},
        "WorkflowA": {gold_workflow},
        "HypoB": {gen_hypo},
        "WorkflowB": {gen_workflow}
        }}
        ```

        {variable_question}"""
dimension_question = """
        Question: Does HypoB exhibit the same relation as HypoA?
        Compare using the following example hierarchy of relationships (based on specificity): \
        "there exists a relationship" > "positive relationship" > "positive AND (linear OR quadratic)" > "positive AND linear."
        Options: A) very similar B) similar but general than HypoA C) different
        Return your answer as a JSON object in the following format:
        ```json
        {{
        "answer": one of the options from A) very similar B) similar but general than HypoA C) different
        "explanation": a short text explanation about the relationship comparison
        }}```
        Answer:"""
\end{minted}
\vspace{-1em}
\caption{Prompt for relationship alignment between two sub-hypotheses.}
\label{list:rel}
\end{listing}

%% file: DiscoveryBench.bbl
\begin{thebibliography}{55}
\providecommand{\natexlab}[1]{#1}
\providecommand{\url}[1]{\texttt{#1}}
\expandafter\ifx\csname urlstyle\endcsname\relax
  \providecommand{\doi}[1]{doi: #1}\else
  \providecommand{\doi}{doi: \begingroup \urlstyle{rm}\Url}\fi

\bibitem[Alexander et~al.(1982)Alexander, Riordan, Fennessey, and Pallas]{alexander1982social}
K.~L. Alexander, C.~Riordan, J.~Fennessey, and A.~M. Pallas.
\newblock Social background, academic resources, and college graduation: Recent evidence from the national longitudinal survey.
\newblock \emph{American Journal of Education}, 90\penalty0 (4):\penalty0 315--333, 1982.

\bibitem[Alves et~al.(2023)Alves, Kalinowski, Giray, Mendez, Lavesson, Azevedo, Villamizar, Escovedo, Lopes, Biffl, et~al.]{alves2023status}
A.~P.~S. Alves, M.~Kalinowski, G.~Giray, D.~Mendez, N.~Lavesson, K.~Azevedo, H.~Villamizar, T.~Escovedo, H.~Lopes, S.~Biffl, et~al.
\newblock Status quo and problems of requirements engineering for machine learning: Results from an international survey.
\newblock In \emph{International Conference on Product-Focused Software Process Improvement}, pages 159--174. Springer, 2023.

\bibitem[Apel and Sweeten(2010)]{apel2010impact}
R.~Apel and G.~Sweeten.
\newblock The impact of incarceration on employment during the transition to adulthood.
\newblock \emph{Social problems}, 57\penalty0 (3):\penalty0 448--479, 2010.

\bibitem[Appiah(2017)]{appiah2017effect}
E.~N. Appiah.
\newblock The effect of education expenditure on per capita gdp in developing countries.
\newblock \emph{International Journal of Economics and Finance}, 9\penalty0 (10):\penalty0 136--144, 2017.

\bibitem[Brinkmann(2019)]{brinkmann2019copper}
J.~Brinkmann.
\newblock Copper output, demand for wood and energy expenditure--evaluating economic aspects of bronze age metallurgy.
\newblock \emph{How’s life}, pages 11--34, 2019.

\bibitem[Brozio et~al.(2019)Brozio, M{\"u}ller, Furholt, Kirleis, Dreibrodt, Feeser, D{\"o}rfler, Weinelt, Raese, and Bock]{brozio2019monuments}
J.~P. Brozio, J.~M{\"u}ller, M.~Furholt, W.~Kirleis, S.~Dreibrodt, I.~Feeser, W.~D{\"o}rfler, M.~Weinelt, H.~Raese, and A.~Bock.
\newblock Monuments and economies: What drove their variability in the middle-holocene neolithic?
\newblock \emph{The Holocene}, 29\penalty0 (10):\penalty0 1558--1571, 2019.

\bibitem[Brozio et~al.(2024)Brozio, Kneisel, Schaefer-Di~Maida, Laabs, Feeser, Ribeiro, and Schultrich]{brozio2024patterns}
J.~P. Brozio, J.~Kneisel, S.~Schaefer-Di~Maida, J.~Laabs, I.~Feeser, A.~Ribeiro, and S.~Schultrich.
\newblock Patterns of socio-economic cultural transformations in neolithic and bronze age societies in the central northern european plain: Human-environmental interaction concerning bourdieu’s forms of capital.
\newblock In \emph{Perspectives on Socio-environmental Transformations in Ancient Europe}, pages 105--142. Springer, 2024.

\bibitem[Cerezer et~al.(2023)Cerezer, Dambros, Coelho, Cassemiro, Barreto, Albert, W{\"u}est, and Graham]{cerezer2023accelerated}
F.~O. Cerezer, C.~S. Dambros, M.~T. Coelho, F.~A. Cassemiro, E.~Barreto, J.~S. Albert, R.~O. W{\"u}est, and C.~H. Graham.
\newblock Accelerated body size evolution in upland environments is correlated with recent speciation in south american freshwater fishes.
\newblock \emph{Nature Communications}, 14\penalty0 (1):\penalty0 6070, 2023.

\bibitem[Chen et~al.(2021)Chen, Tworek, Jun, Yuan, Pinto, Kaplan, Edwards, Burda, Joseph, Brockman, et~al.]{chen2021evaluating}
M.~Chen, J.~Tworek, H.~Jun, Q.~Yuan, H.~P. d.~O. Pinto, J.~Kaplan, H.~Edwards, Y.~Burda, N.~Joseph, G.~Brockman, et~al.
\newblock Evaluating large language models trained on code.
\newblock \emph{arXiv preprint arXiv:2107.03374}, 2021.

\bibitem[Dal~Corso et~al.(2019)Dal~Corso, Kirleis, Kneisel, Taylor, Wieckowska-L{\"u}th, and Zanon]{dal2019s}
M.~Dal~Corso, W.~Kirleis, J.~Kneisel, N.~Taylor, M.~Wieckowska-L{\"u}th, and M.~Zanon.
\newblock \emph{How's Life? Living Conditions in the 2nd and 1st Millennia BCE}.
\newblock Sidestone Press, 2019.

\bibitem[Dougherty(2003)]{dougherty2003numeracy}
C.~Dougherty.
\newblock Numeracy, literacy and earnings: Evidence from the national longitudinal survey of youth.
\newblock \emph{Economics of education review}, 22\penalty0 (5):\penalty0 511--521, 2003.

\bibitem[Feeser et~al.(2019)Feeser, D{\"o}rfler, Kneisel, Hinz, and Dreibrodt]{feeser2019human}
I.~Feeser, W.~D{\"o}rfler, J.~Kneisel, M.~Hinz, and S.~Dreibrodt.
\newblock Human impact and population dynamics in the neolithic and bronze age: Multi-proxy evidence from north-western central europe.
\newblock \emph{The Holocene}, 29\penalty0 (10):\penalty0 1596--1606, 2019.

\bibitem[Feurer et~al.(2015)Feurer, Klein, Eggensperger, Springenberg, Blum, and Hutter]{feurer2015efficient}
M.~Feurer, A.~Klein, K.~Eggensperger, J.~Springenberg, M.~Blum, and F.~Hutter.
\newblock Efficient and robust automated machine learning.
\newblock In \emph{NeurIPS}, 2015.

\bibitem[Fu et~al.(2023)Fu, Ng, Jiang, and Liu]{Fu2023GPTScoreEA}
J.~Fu, S.-K. Ng, Z.~Jiang, and P.~Liu.
\newblock Gptscore: Evaluate as you desire.
\newblock \emph{ArXiv}, abs/2302.04166, 2023.
\newblock URL \url{https://api.semanticscholar.org/CorpusID:256662188}.

\bibitem[Gijsbers et~al.(2022)Gijsbers, Bueno, Coors, LeDell, Poirier, Thomas, Bischl, and Vanschoren]{Gijsbers2022AMLBAA}
P.~Gijsbers, M.~L.~P. Bueno, S.~Coors, E.~LeDell, S.~Poirier, J.~Thomas, B.~Bischl, and J.~Vanschoren.
\newblock Amlb: an automl benchmark.
\newblock \emph{ArXiv}, abs/2207.12560, 2022.
\newblock URL \url{https://api.semanticscholar.org/CorpusID:251066648}.

\bibitem[Glass and Hall(2008)]{glass2008brief}
D.~J. Glass and N.~Hall.
\newblock A brief history of the hypothesis.
\newblock \emph{Cell}, 134\penalty0 (3):\penalty0 378--381, 2008.

\bibitem[Heyard and Held(2024)]{heyard2024meta}
R.~Heyard and L.~Held.
\newblock Meta-regression to explain shrinkage and heterogeneity in large-scale replication projects.
\newblock Technical report, Center for Open Science, 2024.

\bibitem[Jin et~al.(2023)Jin, Chollet, Song, and Hu]{Jin2023AutoKerasAA}
H.~Jin, F.~Chollet, Q.~Song, and X.~Hu.
\newblock Autokeras: An automl library for deep learning.
\newblock \emph{J. Mach. Learn. Res.}, 24:\penalty0 6:1--6:6, 2023.

\bibitem[Jumper et~al.(2021)Jumper, Evans, Pritzel, Green, Figurnov, Ronneberger, Tunyasuvunakool, Bates, {\v{Z}}{\'\i}dek, Potapenko, et~al.]{jumper2021highly}
J.~Jumper, R.~Evans, A.~Pritzel, T.~Green, M.~Figurnov, O.~Ronneberger, K.~Tunyasuvunakool, R.~Bates, A.~{\v{Z}}{\'\i}dek, A.~Potapenko, et~al.
\newblock Highly accurate protein structure prediction with alphafold.
\newblock \emph{Nature}, 596\penalty0 (7873):\penalty0 583--589, 2021.

\bibitem[Kaiser and Schier(2021)]{kaiser2021time}
E.~Kaiser and W.~Schier.
\newblock \emph{Time and Materiality: Periodization and Regional Chronologies at the Transition from Bronze to Iron Age in Eurasia (1200-600 BCE). Edited by Elke Kaiser and Wolfram Schier}.
\newblock Verlag Marie Leidorf GmbH, 2021.

\bibitem[Kitano(2016)]{kitano2016artificial}
H.~Kitano.
\newblock Artificial intelligence to win the nobel prize and beyond: Creating the engine for scientific discovery.
\newblock \emph{AI magazine}, 37\penalty0 (1):\penalty0 39--49, 2016.

\bibitem[Kneisel(2021)]{kneisel2021chronology}
J.~Kneisel.
\newblock Chronology and transformation. the transition from bronze to iron age in northern europe.
\newblock \emph{Time and materiality: Periodization and regional chronologies at the transition from Bronze to Iron Age in Eurasia (1200--600 BCE)}, pages 237--263, 2021.

\bibitem[Langley(1981)]{Langley1981DataDrivenDO}
P.~Langley.
\newblock Data-driven discovery of physical laws.
\newblock \emph{Cogn. Sci.}, 5:\penalty0 31--54, 1981.
\newblock URL \url{https://api.semanticscholar.org/CorpusID:39694251}.

\bibitem[LeDell and Poirier(2020)]{LeDell2020H2OAS}
E.~LeDell and S.~Poirier.
\newblock {H2O AutoML}: Scalable automatic machine learning.
\newblock In \emph{Proceedings of the AutoML Workshop at ICML}, 2020.

\bibitem[Li et~al.(2024)Li, Fox, and Goodman]{Li2024AutomatedSM}
M.~Y. Li, E.~B. Fox, and N.~D. Goodman.
\newblock Automated statistical model discovery with language models.
\newblock \emph{ArXiv}, abs/2402.17879, 2024.
\newblock URL \url{https://api.semanticscholar.org/CorpusID:268041863}.

\bibitem[Li et~al.(2023)Li, Zhang, Dubois, Taori, Gulrajani, Guestrin, Liang, and Hashimoto]{alpaca_eval}
X.~Li, T.~Zhang, Y.~Dubois, R.~Taori, I.~Gulrajani, C.~Guestrin, P.~Liang, and T.~B. Hashimoto.
\newblock Alpacaeval: An automatic evaluator of instruction-following models.
\newblock \url{https://github.com/tatsu-lab/alpaca_eval}, 2023.

\bibitem[Lin et~al.(2024)Lin, Chandu, Brahman, Deng, Ravichander, Pyatkin, Bras, and Choi]{wildbench2024}
B.~Y. Lin, K.~Chandu, F.~Brahman, Y.~Deng, A.~Ravichander, V.~Pyatkin, R.~L. Bras, and Y.~Choi.
\newblock Wildbench: Benchmarking language models with challenging tasks from real users in the wild, 2024.
\newblock URL \url{https://huggingface.co/spaces/allenai/WildBench}.

\bibitem[Liu et~al.(2024)Liu, Wu, Wu, Lu, Chang, and Feng]{liu2024llms}
X.~Liu, Z.~Wu, X.~Wu, P.~Lu, K.-W. Chang, and Y.~Feng.
\newblock Are llms capable of data-based statistical and causal reasoning? benchmarking advanced quantitative reasoning with data.
\newblock \emph{arXiv preprint arXiv:2402.17644}, 2024.

\bibitem[Lorenz(2018)]{lorenz2018kommunikationsstrukturen}
L.~Lorenz.
\newblock \emph{Kommunikationsstrukturen mittelneolithischer Gesellschaften im nordmitteleurop{\"a}ischen Tiefland}.
\newblock Verlag Dr. Rudolf Habelt GmbH, in Kommission, 2018.

\bibitem[Madaan et~al.(2023)Madaan, Tandon, Gupta, Hallinan, Gao, Wiegreffe, Alon, Dziri, Prabhumoye, Yang, Gupta, Majumder, Hermann, Welleck, Yazdanbakhsh, and Clark]{madaan2023selfrefine}
A.~Madaan, N.~Tandon, P.~Gupta, S.~Hallinan, L.~Gao, S.~Wiegreffe, U.~Alon, N.~Dziri, S.~Prabhumoye, Y.~Yang, S.~Gupta, B.~P. Majumder, K.~Hermann, S.~Welleck, A.~Yazdanbakhsh, and P.~Clark.
\newblock Self-refine: Iterative refinement with self-feedback.
\newblock In \emph{Thirty-seventh Conference on Neural Information Processing Systems}, 2023.
\newblock URL \url{https://openreview.net/forum?id=S37hOerQLB}.

\bibitem[Maida et~al.(2023)]{maida2023unter}
S.-D. Maida et~al.
\newblock Unter h{\"u}geln: Bronzezeitliche transformationsprozesse in schleswig-holstein am beispiel des fundplatzes von mang de bargen (bornh{\"o}ved, kr. segeberg) band 1, 2023.

\bibitem[Majumder et~al.(2023)Majumder, Mishra, Jansen, Tafjord, Tandon, Zhang, Callison-Burch, and Clark]{majumder2023clin}
B.~P. Majumder, B.~D. Mishra, P.~Jansen, O.~Tafjord, N.~Tandon, L.~Zhang, C.~Callison-Burch, and P.~Clark.
\newblock {CLIN}: A continually learning language agent for rapid task adaptation and generalization.
\newblock \emph{arXiv preprint arXiv:2310.10134}, 2023.

\bibitem[Majumder et~al.(2024)Majumder, Surana, Agarwal, Hazra, Sabharwal, and Clark]{majumder2024data}
B.~P. Majumder, H.~Surana, D.~Agarwal, S.~Hazra, A.~Sabharwal, and P.~Clark.
\newblock Data-driven discovery with large generative models.
\newblock \emph{ICML}, 2024.

\bibitem[Ottaviano et~al.(2013)Ottaviano, Peri, and Wright]{ottaviano2013immigration}
G.~I.~P. Ottaviano, G.~Peri, and G.~C. Wright.
\newblock Immigration, offshoring, and american jobs.
\newblock \emph{American Economic Review}, 103\penalty0 (5):\penalty0 1925--1959, 2013.

\bibitem[Pal(2023)]{pal2023impact}
L.~C. Pal.
\newblock Impact of education on economic development.
\newblock \emph{Khazanah Pendidikan Islam}, 5\penalty0 (1):\penalty0 10--19, 2023.

\bibitem[Palmisano et~al.(2021)Palmisano, Bevan, Kabelindde, Roberts, and Shennan]{palmisano2021long}
A.~Palmisano, A.~Bevan, A.~Kabelindde, N.~Roberts, and S.~Shennan.
\newblock Long-term demographic trends in prehistoric italy: Climate impacts and regionalised socio-ecological trajectories.
\newblock \emph{Journal of world prehistory}, 34\penalty0 (3):\penalty0 381--432, 2021.

\bibitem[Parkinson et~al.(2021)Parkinson, McLaughlin, Esposito, Stoddart, and Malone]{parkinson2021radiocarbon}
E.~W. Parkinson, T.~R. McLaughlin, C.~Esposito, S.~Stoddart, and C.~Malone.
\newblock Radiocarbon dated trends and central mediterranean prehistory.
\newblock \emph{Journal of world prehistory}, 34:\penalty0 317--379, 2021.

\bibitem[Rambeli et~al.(2021)Rambeli, Marikan, Podivinsky, Amiruddin, and Ismail]{rambeli2021dynamic}
N.~Rambeli, D.~A.~A. Marikan, J.~M. Podivinsky, R.~Amiruddin, and I.~Ismail.
\newblock The dynamic impact of government expenditure in education on economic growth.
\newblock \emph{International Journal of Business and Society}, 22\penalty0 (3):\penalty0 1487--1507, 2021.

\bibitem[Riera et~al.(2024)Riera, Pino, S{\'a}ez, Aymerich, and Melero]{riera2024effect}
M.~Riera, J.~Pino, L.~S{\'a}ez, P.~Aymerich, and Y.~Melero.
\newblock Effect of introduction pathways on the invasion success of non-native plants along environmental gradients.
\newblock \emph{Biological Invasions}, pages 1--20, 2024.

\bibitem[Roziere et~al.(2023)Roziere, Gehring, Gloeckle, Sootla, Gat, Tan, Adi, Liu, Remez, Rapin, et~al.]{roziere2023code}
B.~Roziere, J.~Gehring, F.~Gloeckle, S.~Sootla, I.~Gat, X.~E. Tan, Y.~Adi, J.~Liu, T.~Remez, J.~Rapin, et~al.
\newblock Code llama: Open foundation models for code.
\newblock \emph{arXiv preprint arXiv:2308.12950}, 2023.

\bibitem[Schick et~al.(2024)Schick, Dwivedi-Yu, Dess{\`\i}, Raileanu, Lomeli, Hambro, Zettlemoyer, Cancedda, and Scialom]{schick2024toolformer}
T.~Schick, J.~Dwivedi-Yu, R.~Dess{\`\i}, R.~Raileanu, M.~Lomeli, E.~Hambro, L.~Zettlemoyer, N.~Cancedda, and T.~Scialom.
\newblock Toolformer: Language models can teach themselves to use tools.
\newblock \emph{Advances in Neural Information Processing Systems}, 36, 2024.

\bibitem[Shao et~al.(2023)Shao, Wang, Xu, Wei, Yu, Zhang, Yao, Jin, Cao, Cong, Jensen, and Cheng]{Shao2023ExploringPI}
Z.~Shao, F.~Wang, Y.~Xu, W.~Wei, C.~Yu, Z.~Zhang, D.~Yao, G.~Jin, X.~Cao, G.~Cong, C.~S. Jensen, and X.~Cheng.
\newblock Exploring progress in multivariate time series forecasting: Comprehensive benchmarking and heterogeneity analysis.
\newblock \emph{ArXiv}, abs/2310.06119, 2023.

\bibitem[Shashidhar et~al.(2023)Shashidhar, Chinta, Sahai, Wang, and Ji]{Shashidhar2023DemocratizingLA}
S.~Shashidhar, A.~Chinta, V.~Sahai, Z.~Wang, and H.~Ji.
\newblock Democratizing llms: An exploration of cost-performance trade-offs in self-refined open-source models.
\newblock \emph{ArXiv}, abs/2310.07611, 2023.
\newblock URL \url{https://api.semanticscholar.org/CorpusID:263834891}.

\bibitem[Shinn et~al.(2023)Shinn, Cassano, Labash, Gopinath, Narasimhan, and Yao]{Shinn2023ReflexionLA}
N.~Shinn, F.~Cassano, B.~Labash, A.~Gopinath, K.~Narasimhan, and S.~Yao.
\newblock Reflexion: Language agents with verbal reinforcement learning.
\newblock In \emph{NeurIPS}, 2023.
\newblock URL \url{https://api.semanticscholar.org/CorpusID:258833055}.

\bibitem[Smith et~al.(2005)Smith, Bogin, and Bishai]{smith2005time}
P.~K. Smith, B.~Bogin, and D.~Bishai.
\newblock Are time preference and body mass index associated?: Evidence from the national longitudinal survey of youth.
\newblock \emph{Economics \& Human Biology}, 3\penalty0 (2):\penalty0 259--270, 2005.

\bibitem[Sommerfeld(2013)]{sommerfeld2013gerategeld}
C.~Sommerfeld.
\newblock \emph{Ger{\"a}tegeld Sichel: studien zur monet{\"a}ren Struktur bronzezeitlicher Horte im n{\"o}rdlichen Mitteleuropa}, volume~19.
\newblock Walter de Gruyter, 2013.

\bibitem[Thompson and Skau(2023)]{thompson2023scope}
W.~H. Thompson and S.~Skau.
\newblock On the scope of scientific hypotheses.
\newblock \emph{Royal Society Open Science}, 10\penalty0 (8):\penalty0 230607, 2023.

\bibitem[Weatherly et~al.(2022)Weatherly, Lopez, and Tierra]{weatherly2022impact}
H.~Weatherly, K.~Lopez, and C.~Tierra.
\newblock The impact of education on gdp per capita.
\newblock 2022.

\bibitem[{W}es {M}c{K}inney(2010)]{mckinney-proc-scipy-2010}
{W}es {M}c{K}inney.
\newblock {D}ata {S}tructures for {S}tatistical {C}omputing in {P}ython.
\newblock In {S}t\'efan van~der {W}alt and {J}arrod {M}illman, editors, \emph{{P}roceedings of the 9th {P}ython in {S}cience {C}onference}, pages 56 -- 61, 2010.
\newblock \doi{10.25080/Majora-92bf1922-00a}.

\bibitem[Yang et~al.(2022)Yang, Liu, Li, Wu, Wang, Zhao, and Han]{Yang2022ASL}
Z.~Yang, X.~Liu, T.~Li, D.~Wu, J.~Wang, Y.~Zhao, and H.~Han.
\newblock A systematic literature review of methods and datasets for anomaly-based network intrusion detection.
\newblock \emph{Comput. Secur.}, 116:\penalty0 102675, 2022.

\bibitem[Yao et~al.(2023)Yao, Zhao, Yu, Du, Shafran, Narasimhan, and Cao]{yao2023react}
S.~Yao, J.~Zhao, D.~Yu, N.~Du, I.~Shafran, K.~R. Narasimhan, and Y.~Cao.
\newblock {ReAct}: Synergizing reasoning and acting in language models.
\newblock In \emph{ICLR}, 2023.
\newblock URL \url{https://openreview.net/forum?id=WE_vluYUL-X}.

\bibitem[Yuan et~al.(2023)Yuan, Liu, Zi, Liu, Peng, and Lou]{Yuan2023EvaluatingIL}
Z.~Yuan, J.~Liu, Q.~Zi, M.~Liu, X.~Peng, and Y.~Lou.
\newblock Evaluating instruction-tuned large language models on code comprehension and generation.
\newblock \emph{ArXiv}, abs/2308.01240, 2023.
\newblock URL \url{https://api.semanticscholar.org/CorpusID:260379087}.

\bibitem[Zaw et~al.(2016)Zaw, Hamilton, and Darity]{zaw2016race}
K.~Zaw, D.~Hamilton, and W.~Darity.
\newblock Race, wealth and incarceration: Results from the national longitudinal survey of youth.
\newblock \emph{Race and Social Problems}, 8:\penalty0 103--115, 2016.

\bibitem[Zeng et~al.(2023)Zeng, Yu, Gao, Meng, Goyal, and Chen]{Zeng2023EvaluatingLL}
Z.~Zeng, J.~Yu, T.~Gao, Y.~Meng, T.~Goyal, and D.~Chen.
\newblock Evaluating large language models at evaluating instruction following.
\newblock \emph{ArXiv}, abs/2310.07641, 2023.
\newblock URL \url{https://api.semanticscholar.org/CorpusID:263834884}.

\bibitem[Zhang et~al.(2023)Zhang, Gong, Wu, Liu, and Zhou]{Zhang2023AutoMLGPTAM}
S.~Zhang, C.~Gong, L.~Wu, X.~Liu, and M.~Zhou.
\newblock {AutoML-GPT}: Automatic machine learning with {GPT}.
\newblock \emph{ArXiv}, abs/2305.02499, 2023.

\end{thebibliography}
